\documentclass[logo,copyright]{msas}

\usepackage{csquotes}

\usepackage[backend=bibtex,style=alphabetic,minnames=2,maxnames=5,maxcitenames=2,maxbibnames=99,natbib=true]{biblatex}
\usepackage{amsmath}
\usepackage{tikz}
\usepackage{mathdots}
\usepackage{yhmath}
\usepackage{cancel}
\usepackage{color}
\usepackage{siunitx}
\usepackage{array}
\usepackage{multirow}
\usepackage{amssymb}
\usepackage{gensymb}
\usepackage{tabularx}
\usepackage{nicematrix}
\usepackage{enumitem}
\usepackage{extarrows}
\usepackage{caption}
\usepackage{subcaption}
\usepackage{float}
\usepackage{booktabs}
\usepackage{cuted}
\usepackage{capt-of}
\usepackage[capitalise]{cleveref}
\usetikzlibrary{fadings}
\usetikzlibrary{patterns}
\usetikzlibrary{shadows.blur}
\usetikzlibrary{shapes}

\definecolor{high}{HTML}{ef3b2c}  %
\definecolor{low}{HTML}{fff7bc}  %
\newcommand*{\minval}{0.0}%
\newcommand{\gradientcell}[6]{
    \ifdimcomp{#1pt}{>}{#3 pt}{#1}{
    \ifdimcomp{#1pt}{<}{#2 pt}{#1}{
         \pgfmathparse{int(round(100*(#1/(#3-#2))-(\minval*(100/(#3-#2)))))}
        \xdef\tempa{\pgfmathresult}
        \cellcolor{#5!\tempa!#4!#6} {#1}
    }}
 }
 \newcommand{\rmsegradientz}[1]{
 \gradientcell{#1}{0}{1200}{low}{high}{50}
 }

\newcommand{\rmsegradientt}[1]{
 \gradientcell{#1}{0}{5}{low}{high}{50}
 }
\newcommand{\rmsegradientu}[1]{
 \gradientcell{#1}{0}{6}{low}{high}{50}
 }

\newcommand{\accgradient}[1]{
 \gradientcell{#1}{0}{1}{high}{low}{50}
 }

\crefname{appsec}{Appendix Section}{Appendix Sections}
\crefname{appfig}{Appendix Figure}{Appendix Figures}
\crefname{apptab}{Appendix Table}{Appendix Tables}
\crefname{appeq}{Appendix Equation}{Appendix Equations}

\renewcommand{\thefootnote}{\fnsymbol{footnote}}

\title{ClimaX: \\A foundation model for weather and climate}

\author[1]{Tung Nguyen}  
\author[2]{Johannes Brandstetter}
\author[3]{Ashish Kapoor}
\author[2]{\authorcr Jayesh~K.~Gupta\footnote{Equal contributions as last authors, listed reverse alphabetically}}
  \newcommand\CoAuthorMark{\footnotemark[\arabic{footnote}]} %
\author[1]{Aditya~Grover\protect\CoAuthorMark}

\affil[1]{UCLA}
\affil[2]{Microsoft}
\affil[3]{Scaled Foundations}

\correspondingauthor{tungnd@cs.ucla.edu, johannesb@microsoft.com, ashish.kapoor@gmail.com, \\\qquad \qquad \qquad \qquad jkg@cs.stanford.edu, adityag@cs.ucla.edu}

\begin{abstract}
Most state-of-the-art approaches for weather and climate modeling are based on physics-informed numerical models of the atmosphere. These approaches aim to model the non-linear dynamics and complex interactions between multiple variables, which are challenging to approximate. Additionally, many such numerical models are computationally intensive, especially when modeling the atmospheric phenomenon at a fine-grained spatial and temporal resolution.
Recent data-driven approaches based on machine learning instead aim to directly solve a downstream forecasting or projection task by learning a data-driven functional mapping using deep neural networks. 
However, these networks are trained using curated and homogeneous climate datasets for specific spatiotemporal tasks, and thus lack the generality of numerical models.
We develop and demonstrate ClimaX, a flexible and generalizable deep learning model for weather and climate science that can be trained using heterogeneous datasets spanning different variables, spatio-temporal coverage, and physical groundings.
ClimaX extends the Transformer architecture with novel encoding and aggregation blocks that allow effective use of available compute while maintaining general utility. %
ClimaX is pre-trained with a self-supervised learning objective 
on climate datasets derived from CMIP6. The pre-trained ClimaX can then be fine-tuned to address a breadth of climate and weather tasks, including those that involve atmospheric variables and spatio-temporal scales unseen during pretraining. Compared to existing data-driven baselines, we show that this generality in ClimaX results in superior performance on benchmarks for weather forecasting and climate projections, even when pretrained at lower resolutions and compute budgets. Source code is available at \url{https://github.com/microsoft/ClimaX}.
\end{abstract}

\begin{document}
\maketitle

\begin{figure}[ht]
\centering
    \includegraphics[width=0.67\textwidth]{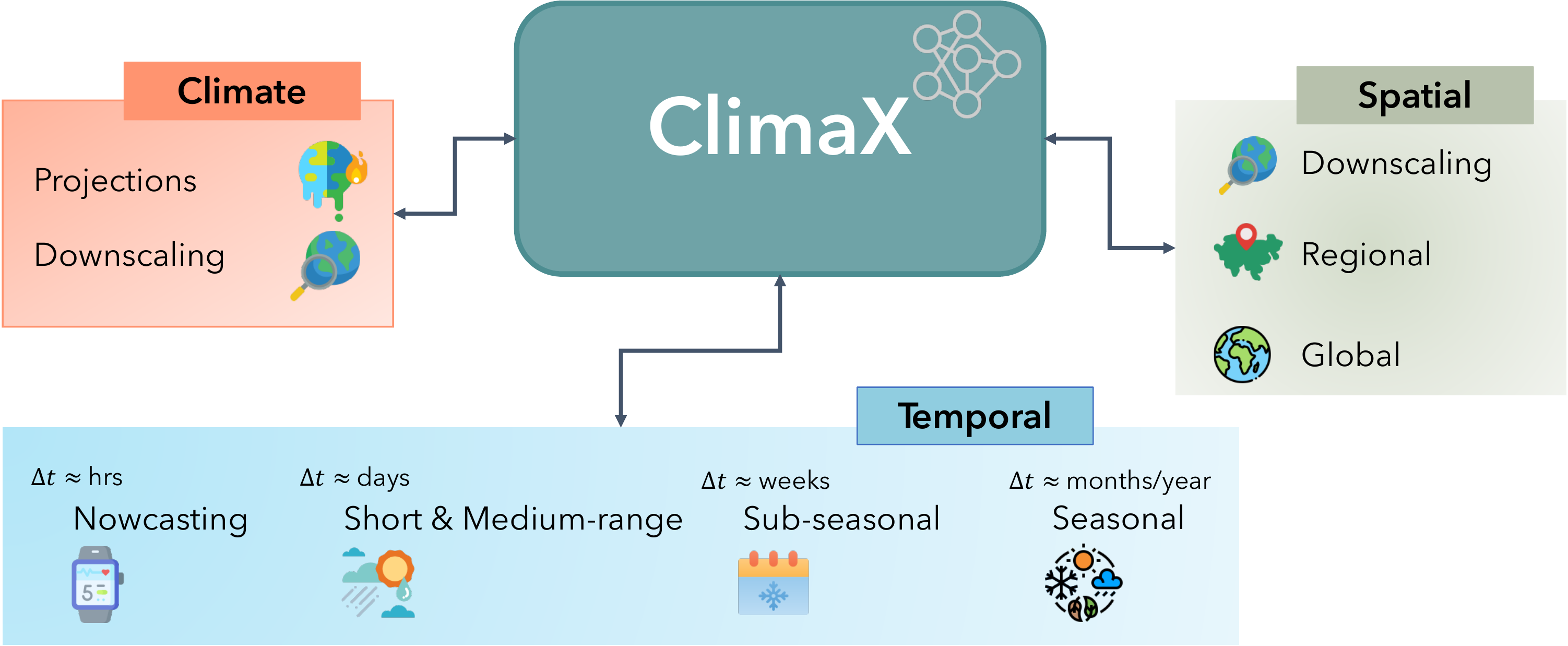}
    \caption{ClimaX is built as a foundation model for any weather and climate modeling task. On the weather front, these tasks include standard forecasting tasks for various lead-time horizons at various resolutions, both globally or regionally. On the climate front, making long term projections and obtaining downscaling results from lower resolution model outputs are standard tasks.\label{fig:climax-feature}
    }
\end{figure}
\begingroup
\small
\hypersetup{linkcolor=black}
\tableofcontents
\endgroup

\clearpage
\setcounter{footnote}{0}
\renewcommand{\thefootnote}{\arabic{footnote}}

\section{Introduction}

Modeling weather and climate is an omnipresent challenge for science and society. With rising concerns around extreme weather events and climate change, there is a growing need for both improved weather forecasts for disaster mitigation and climate projections for  long-term policy making and adaptation efforts~\cite{masson2021climate}. Currently, numerical methods for global modeling of weather and climate are parameterized via various general circulation models (GCM)~\cite{lynch2008origins}. GCMs represent system of differential equations relating the flow of energy and matter in the atmosphere, land, and ocean that can be integrated over time to obtain forecasts for relevant atmospheric variables~\citep{lynch2008origins, bauer2015quiet}.
While extremely useful in practice, GCMs also suffer from many challenges, such as accurately representing physical processes and initial conditions at fine resolutions, as well as technological challenges in large-scale data assimilation and computational simulations~\cite{bauer2020ecmwf}. 
These factors limit their use in many scenarios, especially in simulating atmospheric variables quickly at very short time scales (e.g., a few hours) or accurately at long time scales (e.g., beyond 5-7 days) \cite{zhang2019predictability}.

In contrast, there has been a steady rise in data-driven approaches for  forecasting of atmospheric variables, especially for meteorological applications~\cite{grover2015deep,dueben2018challenges, weber2020deep, scher2019weather, scher2018toward, kashinath2021physics, schultz2021can, reichstein2019deep, huntingford2019machine, schneider2017earth}. 
The key idea here is to train deep neural networks to predict the target atmospheric variables using decades of historical global datasets, such as the ERA-5 reanalysis dataset~\citep{hersbach2020era5}.
Unlike GCMs, these networks are not explicitly grounded in physics, and lack general-purpose utility for Earth system sciences as they are trained for a specific predictive modeling task. Yet, with growing compute and datasets, there is emerging evidence that these models can achieve accuracies competitive with state-of-the-art numerical models in many scenarios, such as nowcasting of precipitation~\citep{ravuri2021skilful, sonderby2020metnet} and medium-range forecasting of variables like temperature, wind and humidity~\citep{weyn2020improving, rasp2021data, keisler2022forecasting, pathak2022fourcastnet, bi2022pangu,lam2022graphcast}.
While these trends are encouraging, there remain concerns regarding the generality of such data-driven methods to diverse real-world scenarios, such as forecasting of extreme weather events and longer-term climate projections, especially under limited spatiotemporal supervision and computational budgets.

Variants of the aforementioned challenges apply broadly throughout machine learning (ML).
In disciplines such as natural language processing and computer vision, it is well acknowledged that ML models trained to solve a single task using supervised learning are label-hungry during training and brittle when deployed outside their training distribution~\cite{taori2020measuring}.  
Recent works have shown that it is possible to mitigate the supervision bottleneck by \textit{pretraining}~\cite{devlin2018bert, he2022masked} large unsupervised ``foundation'' models~\citep{Bommasani2021FoundationModels} on huge passive datasets, such as text and images scraped from the internet~\cite{ramesh2022hierarchical, brown2020language, liu2021swin, reed2022gato}.
Post pretraining, there are many ways to \textit{finetune} the same model on arbitrary target task(s) with little to none (i.e., zero-shot) additional supervision.
Besides low target supervision, these models also generalize better to shifts outside their training distribution~\citep{hendrycks2020pretrained,zhang2022delving}, improving their reliability.

Inspired by the above successes, this work studies the question: how do we design and train a foundation model for weather and climate that can be efficiently adapted for general-purpose tasks concerning the Earth's atmosphere? 
We propose ClimaX, a foundation model for weather and climate. 
For pretraining any foundation model, the key recipe is to train a deep architecture on a large dataset using an unsupervised objective.
For example, many foundation models for language and vision train large transformers on Internet-scale datasets using generative modeling.
While conceptually simple, this scaling recipe is riddled with challenges for weather and climate domains, that we discuss below and propose to resolve with ClimaX.

First, it is unclear what constitutes an Internet-scale passive dataset for pretraining ClimaX.
The size of historical weather and climate datasets at any given time is fixed and increases at an almost constant rate everyday, as it corresponds to processed sensor measurements of naturally occurring phenomena. 
Our first key proposal is to go beyond these datasets to explicitly utilize physics-informed climate simulation models.
Many such models are in use today, for example, the CMIP6 collection~\cite{eyring2016overview} of climate modeling simulations 
consists of runs of $\sim$100 distinct climate models from 49 different climate modeling groups.
We show that the heterogeneity in these simulation datasets serves as a source of rich and plentiful data for pretraining ClimaX.

Second, we need a model architecture that can aptly embrace the heterogeneity of the above climate datasets. 
Climate data is highly multimodal, as observations typically correspond to many different, unbounded variables with varying datatypes (e.g., pressure, temperature, humidity).
Moreover, many observational datasets are irregular in the sense that they differ in their spatiotemporal coverage and might correspond to different subsets of atmospheric variables. 
We resolve the above challenges in ClimaX by repurposing the vision transformer~\cite{dosovitskiy2020image, vaswani2017attention}. 
In contrast to earlier work where the input data is represented as an image with different atmospheric variables treated as the channels thereof~\cite{pathak2022fourcastnet, bi2022pangu}, we treat them as separate modalities to enable more flexible training even with irregular datasets.
This has the side-effect of drastically increasing the sequence length, which we propose to resolve via a cross-attention style channel aggregation scheme prior to the self-attention layers.

Third and last, we need a pretraining objective that can learn complex relationships between the atmospheric variables and permit effective finetuning for downstream tasks.
Given the spatiotemporal nature of climate data, we propose a randomized forecasting objective for pretraining ClimaX. Here, the goal of the model is to forecast an arbitrary set of input variables at an arbitrary time into the future. 
While simple and intuitive, we show that such a pretraining objective aids finetuning to novel tasks and timescales even beyond the pretraining window, such as sub-seasonal to seasonal cumulative predictions, climate projections, and downscaling of climate models.
See Figure~\ref{fig:climax-feature} for a list of tasks considered in this work.

Empirically, we demonstrate that a single pretrained model can be finetuned for many tasks (e.g., multi-scale weather forecasting, climate projections, downscaling) under a range of operating conditions involving different spatiotemporal resolutions, geographical regions, and target prediction variables, including those unseen during training.
Notably, our benchmark results are state-of-the-art on ClimateBench~\cite{watson2022climatebench} and competitive with the operational Integrated Forecasting System (IFS)~\citep{wedi2015modelling} on WeatherBench~\cite{rasp2020weatherbench}, even when our model is trained on moderate resolutions using only a maximum of 80 NVIDIA V100 GPUs.

Finally, we show promising scaling laws of ClimaX
with natural axes of performance improvements for larger number of pre-training datasets,
larger models, and scaling to higher resolution gridded datasets.
While especially the last is in line with recent and concurrent works on data-driven weather forecasting~\cite{pathak2022fourcastnet, bi2022pangu, lam2022graphcast}, 
to the best of our knowledge, ClimaX is the first of its kind data-driven model that can effectively scale using heterogeneous climate datasets during pretraining, and generalize to diverse downstream tasks during finteuning, paving the way for a new generation of data-driven models for Earth systems science.

\section{Background and Related Work}\label{sec:background}

Current weather and climate models in use today rely extensively on numerical methods and computational simulations to predict and understand the Earth's weather and climate systems. 
These tasks include various \emph{numerical weather prediction} (NWP) systems which use computer simulations to make short-term forecasts of weather conditions as well as climate models which use similar techniques to simulate and predict the long-term changes in the Earth's climate.
Most notably, at the core of both weather and climate models lie the same set of primitive equations. 

For climate modeling, earth system models (ESM)~\cite{hurrell2013community}, or ``coupled models'', that couple together simulations which govern the atmosphere, cryosphere, land, and ocean processes are considered the state-of-the-art. Primarily these simulations are based on general circulation models (GCMs)~\cite{satoh2004atmospheric, lynch2008origins, adopted2014climate, masson2021climate} which date back to the works of~\citet{phillips1956general, lorenz1967nature} solving Navier-Stokes equations on a rotation sphere to model fluid circulation. 
These models are often used to perform various \emph{factor sensitivity} studies to examine how the changes in certain forcing factors like greenhouse gas concentrations can affect the global or regional climate and help in \emph{climate projections} to help understand future conditions.

Numerical Weather Prediction (NWP) models share many components of GCMs, especially the atmospheric components~\cite{bauer2015quiet, lynch2008origins, kalnay2003atmospheric}.  
However, incorporating \emph{data assimilation}~\cite{law2015data,grover2022rethinking} which involves combining observations and various measurements of the atmosphere and oceans together with these numerical models is important for accurate forecasts and simulations.
Another significant distinction between weather and climate models is the framing of the solution for underlying equations: \emph{initial value problem} for weather, while \emph{boundary value problem} for climate~\citep{bauer2015quiet}.
Different difficulty levels of these solution approaches results in the fact where climate models tend to be global often at coarser spatio-temporal resolutions while weather models can range from global to local and regional models of very high spatio-temporal resolutions~\citep{warner2010numerical}.

Despite their noted success, including the recent 2021 Nobel Prize in Physics~\cite{ravishankara2022complex}, there is considerable debate around the limitations of general circulation models (GCMs), particularly structural errors across models and the fact that current GCMs are designed to reproduce observed climate~\cite{balaji2022general}.
The climate science community has been aware of these challenges which resulted in the creation of Coupled Model Intercomparison Project (CMIP) as a standardized protocol for evaluating and comparing the performance of different climate models~\citep{meehl2000coupled}.
As we will see in the following sections, not only has CMIP been playing a crucial role in the advancement of our understanding of climate change and its potential impacts, its evaluation procedure has resulted in enormous quantity of data making modern deep learning based approaches quite attractive for many tasks. 
Notably, encoding this knowledge into a ``foundation'' machine learning model with much faster inference and data assimilation capabilities can pave the way for a much wider impact.

\subsection{Data sources}\label{sec:data}
Unlike data in computer vision or natural language processing, weather and climate data is not solely based on sensed data, instead incorporates information from a diverse range of sources. 
For example, \emph{reanalysis} weather data
blends meteorological observations with past short-range
weather forecasts via data assimilation~\cite{bauer2015quiet}.
The data measurements themselves are highly heterogeneous, representing various physical variables with different data types (e.g. pressure, temperature, humidity) that are recorded at different, relatively sparse, spatial locations at different temporal frequencies.
These measurements can be integrated together with known physics inform the design of climate simulations, which again produce data with different variables at different scales. 
From a machine learning perspective, the plethora of available data thus spans multiple axes: from direct weather measurements at land, sea, or atmosphere, over multiple decades of re-analyzed weather data at different spatial scales, to physics-informed climate projections for various scenarios.
Most notably, the data shares the same set of primitive equations, but with fairly different characteristics.
Below we describe two of the most commonly used data sources for weather and climate modeling.

\subsubsection{CMIP6}\label{subsec:data-cmip}
The Coupled Model Intercomparison Project (CMIP)~\citep{meehl2000coupled} is an international effort across different individual climate modeling groups to come together to compare and evaluate their global climate models.
While the main goal of CMIP is to improve the understanding of Earth's climate system and improve the accuracy of its simulations, the recent data from their experimental runs is easily accessible on the CMIP6~\citep{eyring2016overview} archive.
In CMIP6, 
where ``6'' refers to the most recent phase of the project,
$49$ groups are involved with their experiments covering wide range of climate variables including temperature, precipitation, sea level and others from hundreds of models.
This results in global projections of various climate scenarios from as early as 1850 onwards,
all following similar governing equations, but with different \emph{forcings}, e.g., greenhouse gas emissions that affect the climate. 
\subsubsection{ERA5}\label{subsec:data-era5}
The ERA5 reanalysis archive~\cite{hersbach2018era5, hersbach2020era5} 
of the European Center for Medium-Range Weather Forecasting (ECMWF)
is the predominant data source for learning and benchmarking weather forecasting systems.
Once completed, the ERA5 reanalysis is set to embody a detailed record of the global atmosphere, land surface and ocean waves from 1950 onwards.
The currently available ERA5 reanalysis data combines the state of the art forecasting model called Integrated Forecasting System (IFS)~\cite{wedi2015modelling} 
of ECMWF with available observations to provide the best guess of the state of the atmosphere, ocean-wave and land-surface quantities at any point in time.
In its raw form, the available reanalyzed data is huge: 40 years, from 1979 to 2018, on a $0.25\degree \times 0.25\degree$ global latitude-longitude grid of the Earth's sphere, at hourly intervals with different climate variables at 37 different altitude levels plus the Earth's surface. The grid overall contains $721 \times 1440$ grid points for latitude and longitude, respectively. The altitude levels are presented as pressure levels.

\subsection{Tasks}\label{sec:tasks}

Given the scale of data availability, increasing compute requirements of current numerical methods despite it being difficult to incorporate real observational data into them, machine learning is increasingly finding applications in many of the tasks related to weather and climate modeling. 
When it comes to \textbf{weather}, the main task of interest is \emph{forecasting} the future values of key weather variables. 
These tasks can take the following forms depending on temporal and spatial horizons of interest:
\begin{itemize}[align=parleft,left=0pt..1em]
    \item \textbf{Global forecasting} tasks that range from a few hours (i.e., nowcasting) to days and weeks in lead time (i.e., short and medium range forecasting). Often these tasks are evaluated on the ERA5 reanalysis dataset (see \Cref{subsec:data-era5}) with Operational IFS~\citep{wedi2015modelling} 
    of the European Center for Medium-Range Weather Forecasting (ECMWF) being the current state-of-the-art NWP baselines.%
    \item \textbf{Regional forecasting} tasks which could range from weather forecasting in continental North America or Europe to individual state, county or city.
    \item \textbf{Sub-seasonal to seasonal prediction (S2S)}~\cite{vitart2018sub,Vitart2022} which is the task of forecasting the weather with lead times between 2 weeks and 2 months. S2S bridges the gap between weather forecasting and seasonal climate prediction, and is critical to disaster mitigation. Often at such long horizons, predicting instantaneous values of key weather variables can be a difficult task and therefore the focus is often on averaged value of key weather variables over a certain time horizon, e.g.\ weekly average precipitation.  
\end{itemize}
Whereas deep learning approaches for regional or S2S tasks are scarce, most of the recent and concurrent work focuses on global forecasting tasks.
 \citet{rasp2021data} were the first to use pretraining on climate simulations to achieve good data-driven medium-range weather prediction with a ResNet~\cite{he2016deep}, \citet{weyn2020improving}
used CNNs on a cubed sphere for global weather prediction, \citet{weyn2021sub} forecast weather sub-seasonally with a large ensemble of deep-learning weather prediction models, \citet{keisler2022forecasting} applied a graph neural network based approach to weather forecasting,
\citet{ravuri2021skilful} use deep generative models of radar for precipitation nowcasting,
\citet{arcomano2020machine} build a reservoir computing-based, low-resolution, global prediction model, and MetNet~\cite{sonderby2020metnet} takes as input radar and satellite data to forecast probabilistic precipitation maps. 
These approaches are complemented by general machine learning models for fluid dynamics~\cite{li2020fourier, kochkov2021machine, lu2021learning, brandstetter2022clifford, brandstetter2022message}.
Finally, recent state-of-the-art neural weather models such as FourCastNet~\cite{pathak2022fourcastnet}, Pangu-weather~\cite{bi2022pangu}, or GraphCast~\cite{lam2022graphcast}, which also perform global forecasting tasks, use the highest resolution $0.25\degree$ ERA5 data, and are optimized on the respective hardware resources.

On the other hand, \textbf{climate} tasks have to deal with much longer time horizons.
Possible categories of tasks where machine learning can help include climate projection and climate model downscaling:
\begin{itemize}[align=parleft,left=0pt..1em]
\item \textbf{Climate projection} is the task of 
generating estimates of climate change under different future socio-economic scenarios.
Usually, this takes the form of figuring out the response of the climate system to different forcing factors such as greenhouse gases and aerosol emissions. 
Climate projection is a crucial task in understanding and preparing for the potential impacts of climate change.

While the application of machine learning in this field is still in its early stages, recent efforts have been made to standardize evaluation in this domain. One example of this is ClimateBench~\citep{watson2022climatebench}, which is a benchmark dataset drawing on CMIP6 to provide an evaluation framework for machine learning models that aim to improve the accuracy of climate projections. This benchmark aims to provide a consistent and reliable evaluation method for various machine learning models that are applied to climate projections.%
\item A more popular application of ideas in machine learning is towards \textbf{downscaling} of climate model. 
Global climate models typically have a coarse spatial resolution, which means that they can only provide a rough estimate of climate conditions at a local or regional scale.
Moreover, the simulations often reflect systematic biases that deviate from trends in the observation data.
The aim of climate model downscaling is to create locally accurate climate information from global climate projections by relating those to observed local climatological conditions.
This process improves the spatial and temporal resolution of the data, making it more suitable for use in local and regional analyses.
Downscaling methods can be divided into \textit{dynamic} approaches that relate outputs of global climate models with those of regional climate models, and \textit{statistical} approaches that infer the desired transformations using data-driven approaches~\cite{wilby1997downscaling}.
Dynamic approaches are physically consistent, but can be slow and have large biases, whereas statistical approaches need large amounts of data to learn expressive mappings that hold for target output scenarios. 
\end{itemize}

Similar to weather forecasting, deep learning has emerged as appealing alternative in climate science as well. Recent approaches comprise surrogate models to emulate climate projections
~\cite{weber2020deep, scher2019weather, scher2018toward, beusch2020emulating, mansfield2020predicting}, extract contextual cues from existing datasets or simulations~\cite{reichstein2019deep, huntingford2019machine, schneider2017earth},
and perform climate model downscaling~\cite{sachindra2018statistical, vandal2017deepsd, bano2020configuration}.
Climate model downscaling usually inputs low-resolution reanalysis data and local orographic information to obtain high-resolution local information.
Many recent approaches are based on convolutional architectures~\cite{hohlein2020comparative, vaughan2021convolutional, markou2022practical}.

\subsection{Foundation models}\label{subsec:foundmodels}
\citet{Bommasani2021FoundationModels} gave the term ``foundation models'' to the emerging paradigm of training scalable deep learning models on broad data via self-supervision which could then be adapted (often via finetuning) to a wide range of downstream tasks. Current notable examples include BERT~\citep{devlin2018bert}, GPT~\citep{brown2020language} and PaLM~\citep{chowdhery2022palm}, in language, CLIP~\citep{radford2021learning}, Florence~\citep{yuan2021florence}, BEiT~\citep{wang2022image} for vision-language. 
Outside applications on data crawled from web, this paradigm has also started finding success in various scientific domains like protein design~\citep{verkuil2022language}. 
Key significance of such models has been identified as \emph{emergence} with respect to model capabilities and \emph{homogenization} with respect to methodologies for different tasks, domains, and modalities, enabled by the principles of transfer learning~\citep{thrun2012learning} at scale.  
While a foundation model itself should be considered incomplete, it can provide a common basis from which various task-specific models can be derived.
Current research at the intersection of weather and climate science and ML has largely focused on designing separate models for every task of interest despite potential availability of fairly diverse large scale data with shared underlying physics and geology across these tasks.
A few recent works have proposed pretraining techniques for satellite imagery and remote sensing~\cite{yuan2020self,cong2022satmae,reed2022scale} but they have so far not been applied to multi-sensory data and variables in weather and climate.

\section{Approach}

Given the availability of large scale data sources, together with shared physics and geology between various weather and climate tasks, we aim to build a generalizable deep learning foundation model. The model needs to be able to input heterogeneous datasets of different variables, and provide spatio-temporal coverage based on physical groundings. We, therefore, first take a closer look at input representations, and next design a model to cope with their heterogeneity - local, global, and across variables.

\subsection{Input representation} \label{sec:input-repr}
We are interested in gridded prediction tasks, in which the model takes an input of shape $V \times H \times W$ and predicts an output of shape $V' \times H' \times W'$. $V$ refers to the number of input variables, which can be weather conditions such as geopotential and temperature, or climate forcing factors such as CO$_2$ and SO$_2$. $H$ and $W$ refer to the spatial resolution of the input data, which depends on how densely we grid the globe. 
 This general representation captures a broad variety of downstream tasks in Earth systems science.
 Similarly, $V', H', W'$ refer to the variables and spatial resolution of the predicted outputs.
We mainly work with two spatial resolutions: $5.625\degree$ ($32 \times 64$ grid points) and $1.40625\degree$ ($128 \times 256$ grid points). Semantically, a $H \times W$ map can represent the entire globe or a specific region such as North America.

\subsection{Model architecture} \label{sec:model_arc}
We aim to design a foundation model that we can pretrain on heterogeneous data sources and then finetune to solve various downstream weather and climate tasks. From \Cref{sec:input-repr}, one could think of the tasks as image-to-image translation problems with $V$ input channels and $V'$ output channels. This makes any image architecture a natural fit, such as UNet~\citep{ronneberger2015u}, ResNet~\citep{he2016deep}, or Vision Transformers (ViT)~\citep{dosovitskiy2020image}. 
However, the settings of climate and weather tasks are much broader, where we may want to make predictions for regional or even spatially incomplete data, forecast unseen climate variables, or finetune the model on data at different resolutions from pretraining. Current CNN-based architectures are not applicable in these scenarios, as they require the input to be perfectly gridded, contain a fixed set of variables, and have a fixed spatial resolution. Transformers-based architectures, on the other hand, provide much better flexibility by treating the image-like data as a set of tokens. Therefore, we build ClimaX architecture upon Vision Transformers (ViT)~\citep{dosovitskiy2020image, vaswani2017attention}, and propose two major architectural changes, namely \emph{variable tokenization} and \emph{variable aggregation} to further improve the flexibility and generality, which we will describe next.

\begin{figure*}
    \centering
    \includegraphics[width=0.99\textwidth]{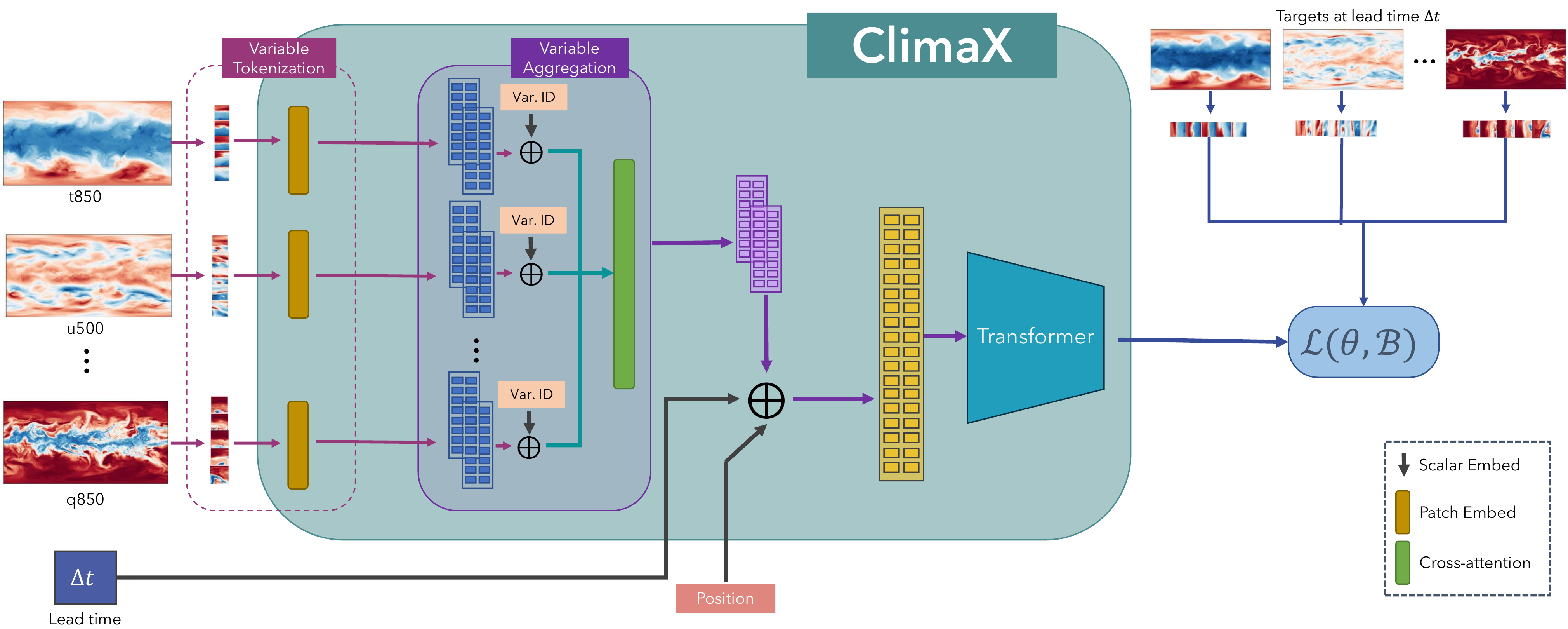}
    \caption{Pretraining phase of ClimaX. 
    Variables are encoded using variable-separate tokenization, and subsequently aggregated using variable aggregation. Together with position embedding and lead time embedding those are fed to the ViT backbone. 
    }
    \label{fig:pretrain}
\end{figure*}

\subsubsection{Variable tokenization}
\begin{figure}[h]
    \centering
    \includegraphics[width=0.9\textwidth]{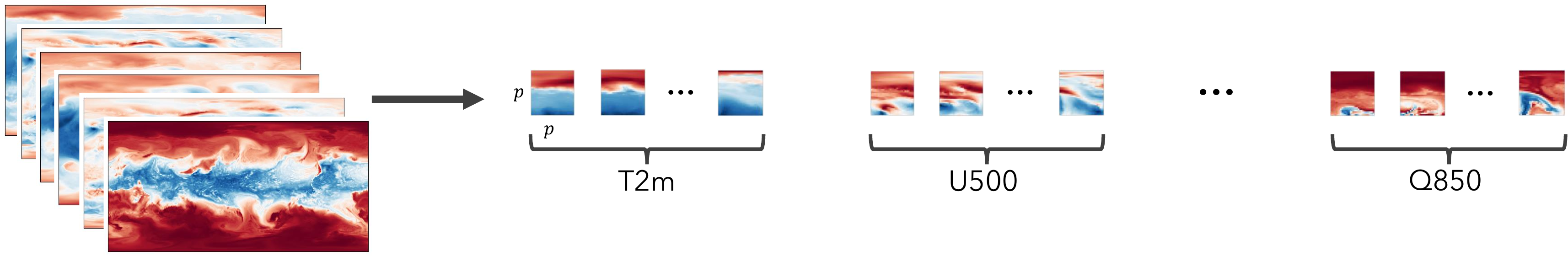}
    \caption{Variable tokenization. Each variable is independently tokenized.
    }
    \label{fig:tokenization}
\end{figure}

Given an input of shape $V \times H \times W$, ViT tokenizes the input into a sequence of $(H/p) \times (W/p) = h \times w$ patches, with each patch having a size of $V \times p^2$, where $p$ is the patch size. This tokenization scheme works well for image data, as $V$ is always the RGB channels, which is the same for all datasets. However, this is not true for climate and weather data, where the number of physical variables can vary between different datasets.
For example, in the CMIP6 project~\citep{eyring2016overview}, each dataset contains simulated data of a different climate model, and thus has a different set of underlying variables. Therefore, we propose \emph{variable tokenization}, a novel tokenization scheme that tokenizes each variable in the input separately. Specifically, each input variable as a spatial map of shape $H \times W$ is tokenized into a sequence of $h \times w$ patches, which results in $V \times h \times w$ patches in total. Finally, each input patch of size $p^2$ is linearly embedded to a vector of dimension $D$, where $D$ is the chosen embedding size. The output of the variable tokenization module therefore has a dimension of $V \times h \times w \times D$.
Figure~\ref{fig:tokenization} illustrates our proposed tokenization scheme.

\subsubsection{Variable aggregation}
\begin{figure}[ht]
    \centering
    \includegraphics[width=0.67\textwidth]{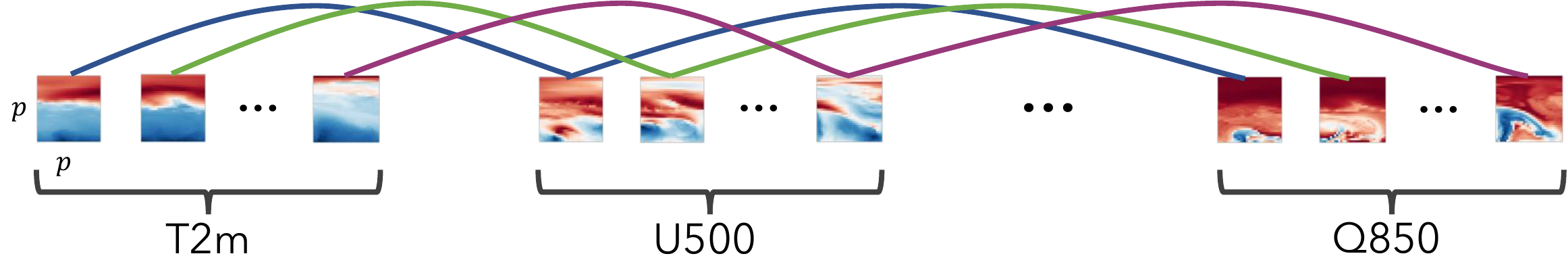}
    \caption{Position-based variable aggregation reduces a sequence of length $V \times h \times w$ to $h \times w$.}
    \label{fig:variable-agg}
\end{figure}

While variable tokenization allows ClimaX to learn from datasets with varying numbers of input variables, it has two inherent problems. First, it results in a sequence of length $V \times h \times w$ which increases linearly with the number of variables. Since we use attention to model the sequence, the memory complexity scales quadratically with the number of variables. This is computationally expensive, as we can have up to $48$ input variables in our experiments. Moreover, because we tokenize each variable separately, the input sequence will contain tokens of different variables with very different physical groundings,
which can create difficulties for the attention layers to learn from. 
We therefore propose \emph{variable aggregation} to solve the two mentioned challenges. For each spatial position in the $h \times w$ map, we perform a cross-attention operation, in which the query is a learnable vector, and the keys and values are the $V$ embedding vectors of $V$ variables at that position. The cross-attention module outputs a single vector for each spatial position, thus reducing the sequence length to $h \times w$, significantly lowering the computational cost. Moreover, the sequence now contains unified tokens with universal semantics, creating an easier task for the attention layers. Figure~\ref{fig:variable-agg} shows our proposed variable aggregation.

\subsubsection{Transformer}
Post variable aggregation, we need a sequence model for generating the output tokens. 
While in principle, one could use any general sequence model, we propose to extend a standard Vision Transformer (ViT).
Moreover, since the standard ViT treats image modeling as pure sequence-to-sequence problems, it can perform tasks that some other variations cannot~\citep{liu2021swin,liu2021swinv2}, such as learning from spatially incomplete data, where the input does not necessarily form a complete grid. This is useful in the regional forecasting task we consider in \Cref{sec:regional-forecast}.
In the experiments, we report results with $8$ attention layers, an embedding size of $1024$, and a hidden dimension of $1024 \times 4$. After the attention layers, we employ a prediction head that takes a token and outputs a vector of size $V' \times p^2$. The prediction head is a 2-layer MLP with a hidden dimension of $1024$. We provide more details in \cref{sec:app:arch}.

\subsection{Datasets}

\subsubsection{Pretraining}
We believe that CMIP6's diversity and scale presents an attractive opportunity for pretraining large-scale foundation models. 
However, handling the inconsistent set of variables across different data sources can be a challenge.
In this work we only use a subset of variables from five different data sources (MPI-ESM, TaiESM, AWI-ESM, HAMMOZ, CMCC) containing global projections of climate scenarios from 1850 to 2015 with the time delta of $6$ hours as described in \cref{tab:cmip6_data}. 
Due to variable original resolution, we choose to simplify our data-loading by regridding them to commonly used resolutions~\citep{rasp2020weatherbench,rasp2021data} of $5.625\degree$ ($32 \times 64$ grid points) and $1.40625\degree$ ($128 \times 256$ grid points)\footnote{Regridding was done using the xesmf Python package~\citep{zhuang2018xesmf} using bilinear interpolation.}.

\subsubsection{Finetuning and evaluation}
We use the ERA5 reanalysis data as described in \cref{sec:app:era5_details}, as the source of datasets for finetuning and evaluation for various weather related downstream tasks.
Due to its large size, it is common to regrid~\citep{rasp2020weatherbench,rasp2021data} the high-resolution data to lower resolutions like $5.625\degree$ ($32 \times 64$ grid points) and $1.40625\degree$ ($128 \times 256$ grid points) to fit within the available computational constraints\footnote{Regridding was done using the xesmf Python package~\citep{zhuang2018xesmf} using bilinear interpolation.}.
We follow the evaluation procedure by~\citet{rasp2021data} and use this data to assess the forecasting performance of our ML models at different lead time horizons.
More details about the individual datasets are in their appropriate experiment sections.

\subsection{Training}

\subsubsection{Pretraining}
We pretrain ClimaX on CMIP6 data to predict future weather conditions given the current conditions. That is, given the weather snapshot $X_t$ of shape $V \times H \times W$ at a particular time $t$, ClimaX learns to predict the future weather scenario $X_{t + \Delta t}$ of the same shape at lead time $\Delta t$. To obtain a pretrained model that is generally applicable to various temporal forecasting tasks, we randomize the lead time from $6$ hours to $168$ hours (i.e., 1 week) during pretraining. We add the lead time embedding to the tokens to inform the model of how long it is forecasting into the future. The lead time embedding module is a single-layer MLP that maps a scalar to a vector of the embedding size $D$. \Cref{fig:pretrain} depicts the forward pass of ClimaX in pretraining. For an input $X_t$, we sample a lead time $\Delta t \sim \mathcal{U}[6, 168]$ and get the corresponding ground truth $X_{t + \Delta t}$. Input variables are tokenized separately using variable tokenization, and are subsequently aggregated at each spatial location, resulting in a sequence of $h \times w$ unified tokens. We add the tokens with the lead time embedding and positional embedding before feeding the sequence to the ViT backbone. The output of the last attention layer is fed to a prediction head, which transforms the sequence back to the original shape of $V \times H \times W$.

We employ the latitude-weighted mean squared error~\citep{rasp2020weatherbench} as our objective function. Given the prediction $\Tilde{X}_{t + \Delta t}$ and the ground truth $X_{t + \Delta t}$, the loss is computed as:
\begin{equation}
    \mathcal{L} = \frac{1}{V \times H \times W} \sum_{v=1}^V \sum_{i=1}^H \sum_{j=1}^W L(i)(\Tilde{X}_{t + \Delta t}^{v,i,j} - X_{t + \Delta t}^{v,i,j})^2, \label{eq:lat_mse}
\end{equation}
in which $L(i)$ is the latitude weighting factor:
\begin{equation}
    L(i) = \frac{\cos(\text{lat}(i))}{\frac{1}{H} \sum_{i'=1}^H \cos(\text{lat}(i'))},
\end{equation}
where $\text{lat}(i)$ is the latitude of the corresponding $i\text{th}$ row of the grid. The latitude weighting term accounts for the non-uniformity in areas when we grid the round globe. Grid cells toward the equator have larger areas than the cells near the pole, and thus should be assigned more weights.

\subsubsection{Finetuning}

ClimaX has four learnable components, including the token embedding layers, the variable aggregation module, the attention blocks, and the prediction head. We evaluate the performance of ClimaX on various downstream tasks, which we categorize into two finetuning scenarios: one in which the downstream variables belong to the set of pretraining variables, and the other with variables unseen during pretraining. In the first case, we finetune the entire model, and in the latter, we replace the embedding layers and the prediction head with newly initialized networks, and either finetune or freeze the other two components. We present more details of each downstream task in \Cref{sec:exps}.

\section{Experiments} \label{sec:exps}
We finetune ClimaX on a diverse set of downstream tasks to evaluate its performance and generality. We categorize the tasks into forecasting, climate projection, and climate downscaling. The experiments aim to answer the following questions:
\begin{itemize}
    \item How does ClimaX perform on global forecasting compared to the current state-of-the-art NWP system?
    \item Can we finetune ClimaX to make forecasts for a specific region or at different temporal horizons from pretraining?
    \item How well does ClimaX perform on climate tasks that are completely different from pretraining?
\end{itemize}
In addition to the main experiments, we analyze the scaling property of ClimaX, i.e., how the performance of ClimaX improves with increasing data size, model capacity, and data resolution. Finally, we perform comprehensive ablation studies to understand the trade-off between computation and performance when finetuning ClimaX.

\subsection{Neural baselines}

In global forecasting, we compare ClimaX with IFS~\citep{wedi2015modelling}, the current gold standard in weather forecasting. In tasks we do not have a baseline, we compare with UNet \citep{ronneberger2015u,gupta2022towards} and ResNet~\citep{he2016deep}, two CNN baselines commonly used in vision tasks. We borrow the ResNet architecture from Weatherbench~\citep{rasp2020weatherbench}. The exact architectural details of these baselines are in \Cref{sec:app:cnn_arc}.

\subsection{Forecasting}

\subsubsection{Global forecasting} \label{sec:global-forecast}
Given global weather conditions $X_t$ at a particular time $t$, we want to forecast the weather at a future time $X_{t + \Delta t}$, in which $\Delta t$ is the lead time. The input variables include $6$ atmospheric variables at $7$ vertical levels, $3$ surface variables, and $3$ constant fields, resulting in $48$ input variables in total. The details of the variables are in \Cref{tab:era5_data}. We evaluate ClimaX on predicting four target variables: geopotential at $500$hPa (Z500), the temperature at $850$hPa (T850), the temperature at $2$ meters from the ground (T2m), and zonal wind speed at $10$ meters from the ground (U10). Z500 and T850 are the two standard verification variables for most medium-range NWP models and are often used for benchmarking in previous deep learning works, while the two surface variables, T2m and U10, are relevant to human activities. We consider seven lead times: $6$ hours, $\{1, 3, 5, 7\}$ days, 2 weeks, and 1 month, which range from nowcasting to short and medium-range forecasting and beyond. We consider predicting each target variable at each lead time a separate task, and finetune a separate model for each task. We discuss alternative finetuning protocols in Section~\ref{sec:ablation}.

We compare ClimaX with IFS and the two CNN baselines on the ERA5 dataset at both $5.625\degree$ and $1.40625\degree$ resolutions. Following~\citep{rasp2020weatherbench}, we split the data into three sets, in which the training data is from $1979$ to $2015$, the validation data is in $2016$, and the test data is in $2017$ and $2018$. We finetune ClimaX and train the other deep learning baselines using the latitude-weighted MSE loss in \Cref{eq:lat_mse}. We perform early stopping on the validation loss for all deep learning models, and evaluate the best checkpoint on the test set. For IFS, we download the predictions from the TIGGE archive~\citep{bougeault2010thorpex} for the year $2018$\footnote{We were not able to download IFS predictions for $2017$ due to some server issues.}. We compare all methods on latitude-weighted root mean squared error (RMSE) and latitude-weighted anomaly correlation coefficient (ACC), two commonly used metrics in previous works. The formulations of the two metrics are in \Cref{sec:app:metrics}. Lower RMSE and  higher ACC indicates better performance. 

\begin{figure}[t]
    \centering
    \includegraphics[width=\textwidth]{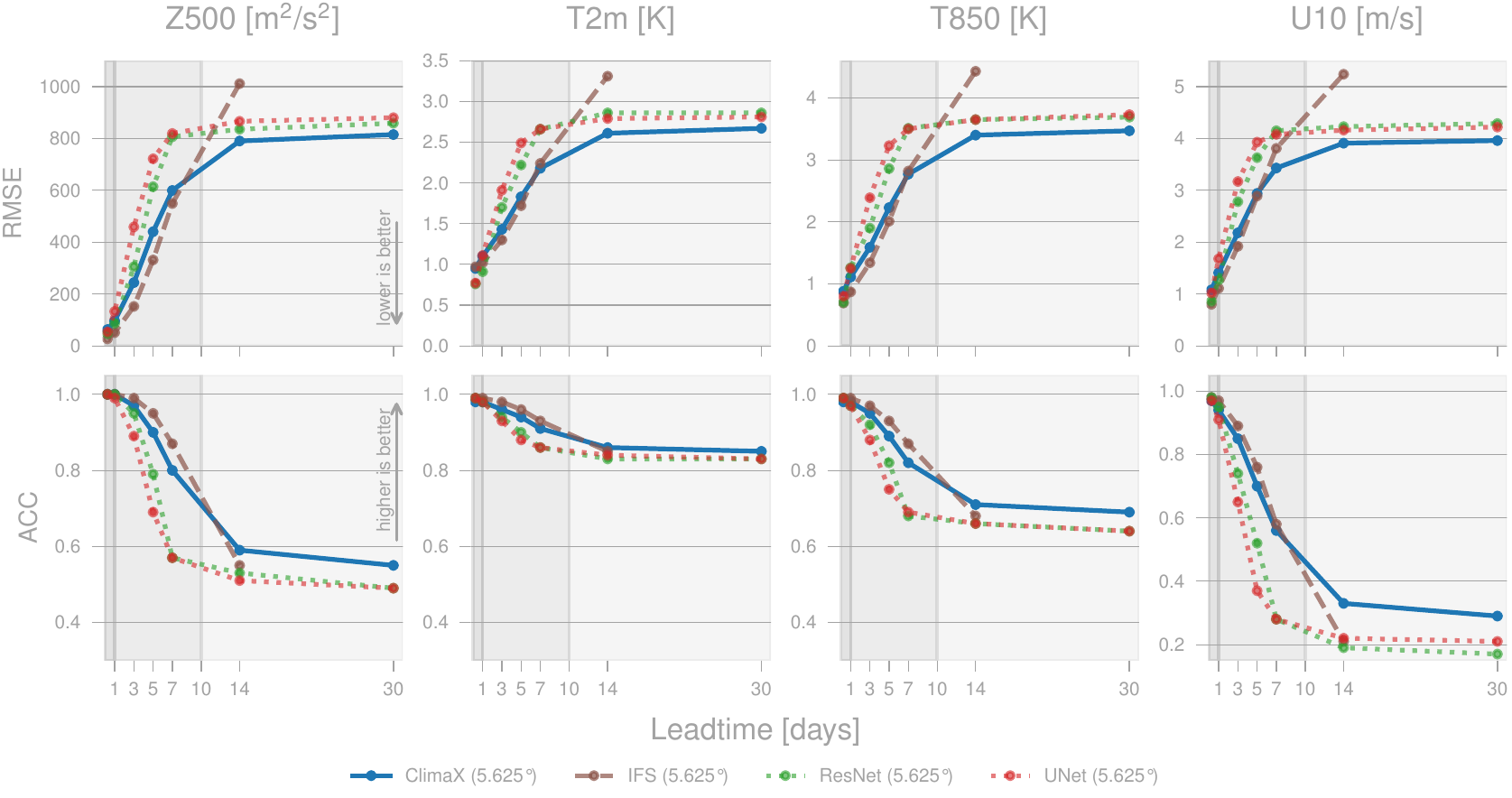}
    \caption{Performance on global forecasting on ERA5 at $5.625\degree$.}\label{fig:global-forecasting-lowres}
\end{figure}

\begin{figure}[t!]
    \centering
    \includegraphics[width=\textwidth]{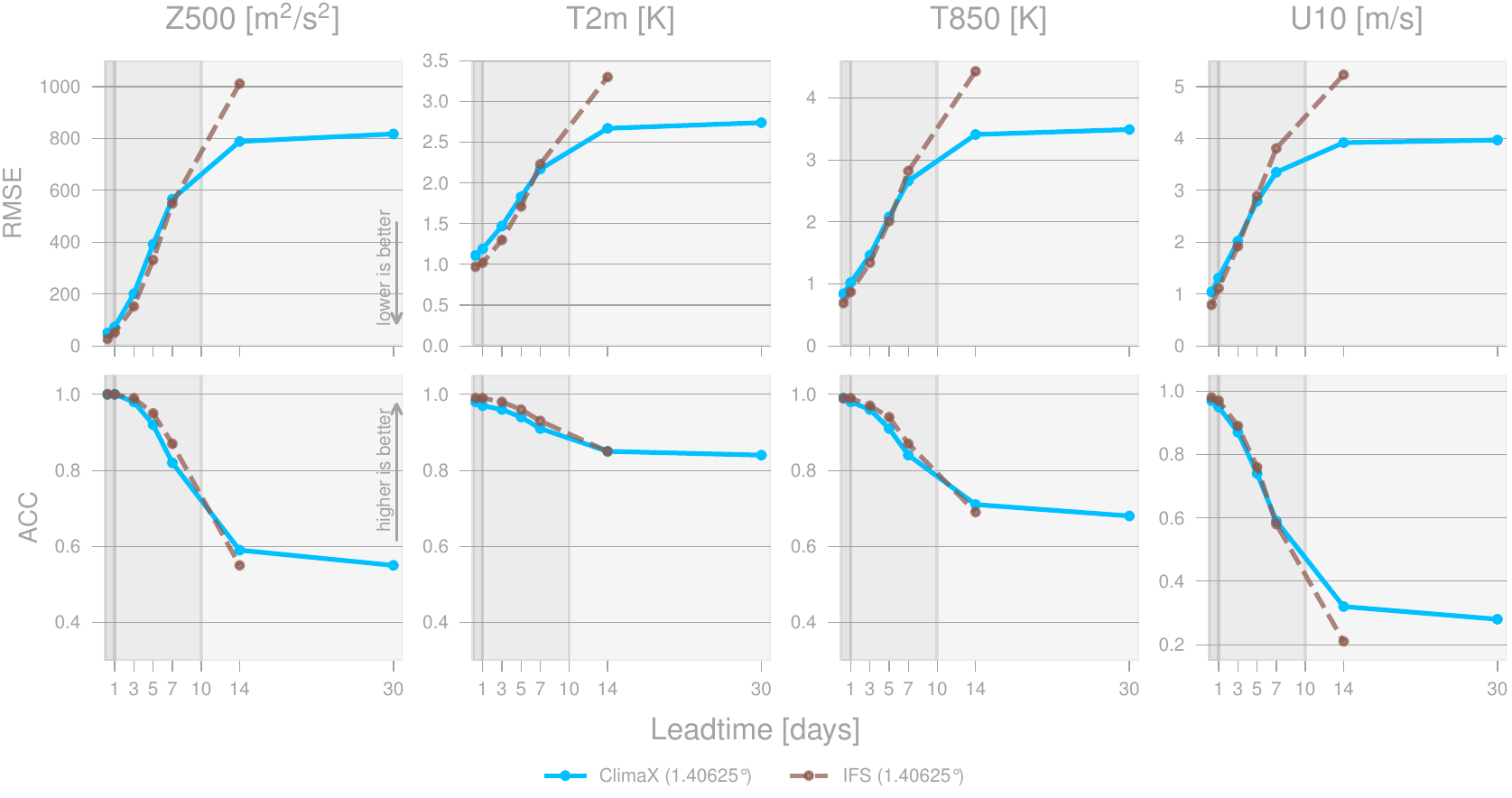}
    \caption{Performance on global forecasting on ERA5 at $1.40625\degree$. 
    }
    \label{fig:global-forecasting}
\end{figure}

\Cref{fig:global-forecasting-lowres,fig:global-forecasting} show the performance of ClimaX and the baselines at $5.625\degree$ and $1.40625\degree$, respectively. At low resolution, IFS outperforms ClimaX on 6-hour to 5-day prediction tasks. On longer horizons, however, ClimaX performs comparably to or slightly better than IFS, especially on 14-day prediction. At higher resolution, the performance of ClimaX closely matches that of IFS even for short horizons, and is superior in forecasting at $7$ days and beyond. The trends are similar for both RMSE and ACC. The two CNN baselines perform similarly and achieve reasonable performance, but lag behind ClimaX and IFS on all tasks.
 We include other additional task-specific baselines~\citep{pathak2022fourcastnet,bi2022pangu,lam2022graphcast} in \Cref{app:results_summary}.
These baselines are trained on higher-resolution ERA5 ($0.25\degree$) so are not directly comparable.


\subsubsection{Regional forecasting} \label{sec:regional-forecast}
\begin{figure}[t]
    \centering
    \includegraphics[width=0.65\textwidth]{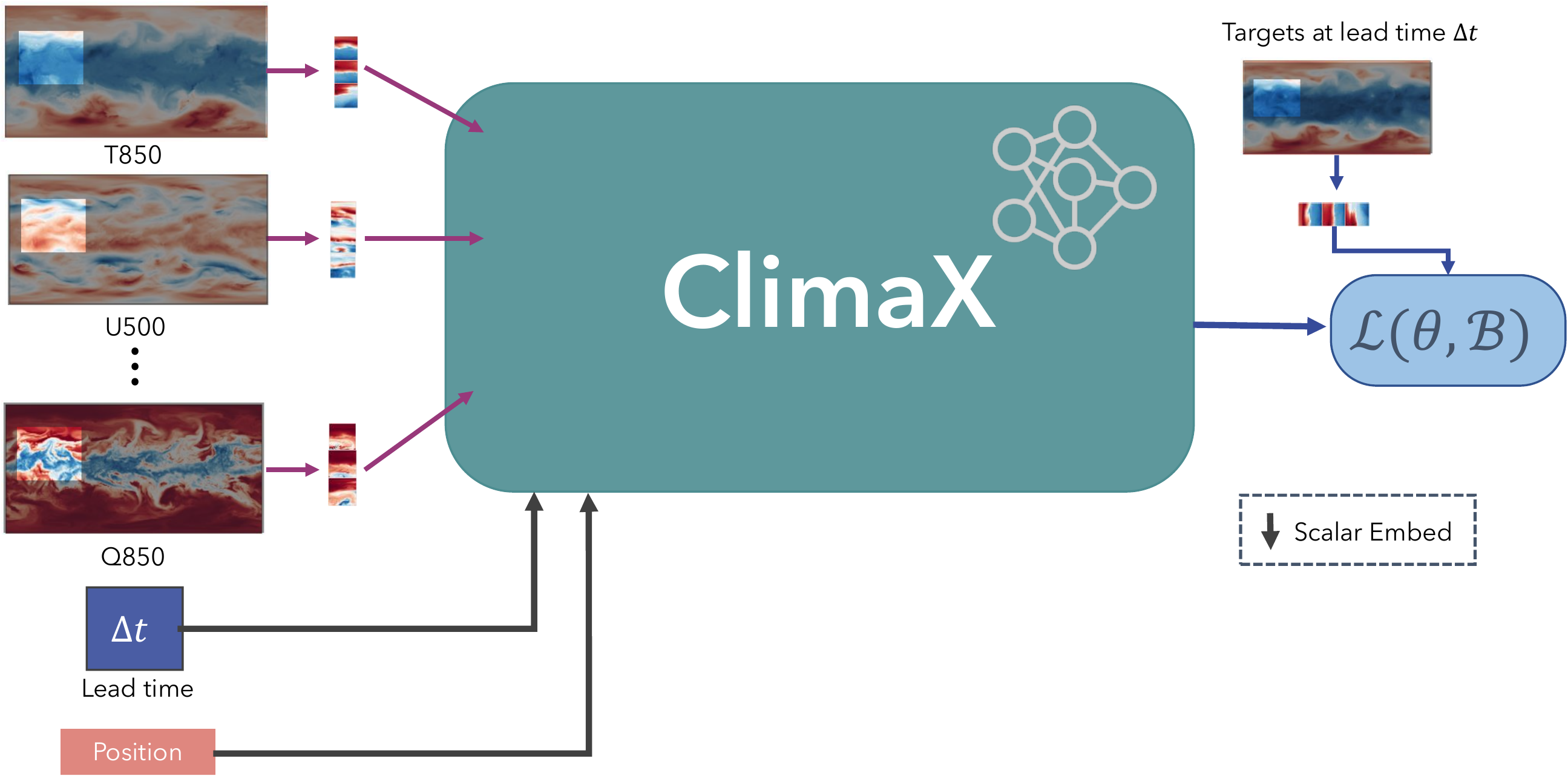}
    \caption{Finetuning setup for Regional Forecasting in North America. 
    }
    \label{fig:regional-finetuning}
\end{figure}

It is not always possible to make global predictions, especially when we only have access to regional data In this section, we evaluate ClimaX on \emph{regional forecasting} of the relevant variables in North America, where the task is to forecast the future weather in North America given the current weather condition in the same region. 
We create a new dataset from the ERA5 data at $1.40625\degree$ that has the same set of variables but just focuses on the North America region. We call this dataset ERA5-NA and present details of how to construct it in \Cref{sec:app:era5_details}. Training, validation, and test splits are done similarly to \Cref{sec:global-forecast}.
\Cref{fig:regional-finetuning} illustrates the finetuning process of ClimaX on this task, where the only difference from global forecasting is the input now only contains tokens that belong to North America. 

Since the task has not been considered in previous works, we compare ClimaX with the two CNN baselines ResNet and UNet, and the scratch-trained version of ClimaX, which we refer to as Cli-ViT. 
In addition, we finetune two ClimaX models, in which one was pretrained on CMIP6 at $1.40625\degree$, and the other was pretrained on $5.625\degree$ data. To finetune the low-resolution model on higher-resolution data, we follow the common practice of interpolating the positional embedding~\citep{dosovitskiy2020image,touvron2021training}. We denote this model as ClimaX-pos-interp.
We evaluate all methods on predicting Z500, T2m, and T850 at lead times of $3$, $5$, and $7$ days. Latitude-weighted RMSE is used as the evaluation metric.

\begin{figure}
    \centering
    \includegraphics{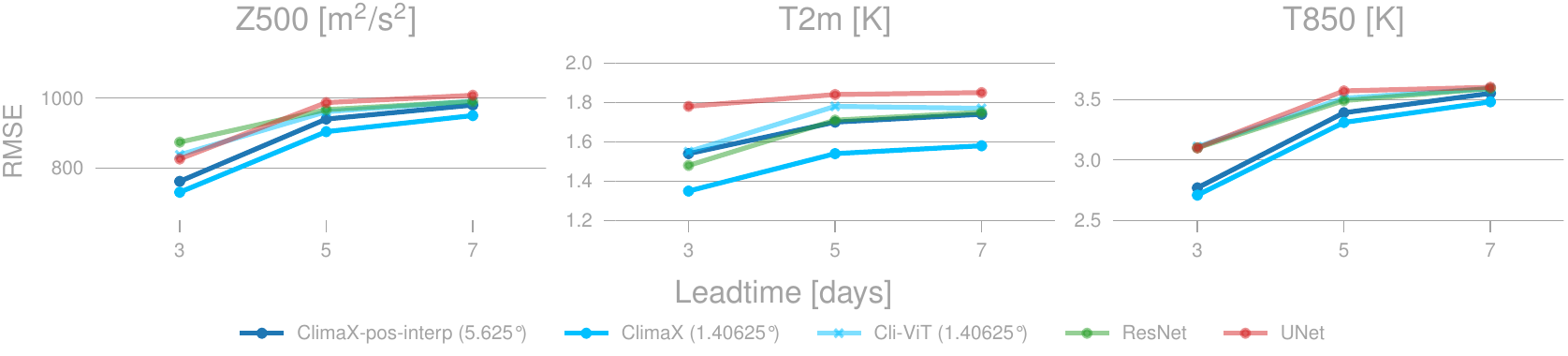}
    \caption{Performance on Regional (North America) forecasting for key variables.}
    \label{fig:northamerica-forecasting}
\end{figure}

\Cref{fig:northamerica-forecasting} compares the performance of ClimaX and the baselines. ClimaX is the best performing method among different target variables and lead times. Interestingly, even though pretrained on data at a lower resolution, ClimaX-pos-interp achieves the second best performance in predicting Z500 and T850, and only underperforms ResNet in predicting T2m at 3-day lead time. This result shows that ClimaX can gain strong performance on tasks that have different spatial coverage or even different spatial resolution from pretraining. 

\subsubsection{Sub-seasonal to seasonal cumulative prediction}
Sub-seasonal to seasonal (S2S) prediction is the task of forecasting at a time range between $2$ weeks and $2$ months~\citep{vitart2018sub}, which bridges the gap between weather forecasting and climate projection. Compared to the other two well-established tasks, S2S prediction has received much less attention, despite having a significant socioeconomic value in disaster mitigation. 
Recent works have proposed data-driven approaches based on both traditional machine learning~\citep{hwang2019improving,prokhorenkova2018catboost,taylor2018forecasting} and deep learning~\citep{weyn2021sub,zhou2021informer,oreshkin2019n}, but their performances often lag behind adaptive bias correction methods~\citep{mouatadid2023adaptive} on standard benchmarks~\citep{mouatadid2023subseasonalclimateusa}.
Here, following the S2S competition (\href{https://s2s-ai-challenge.github.io/}{https://s2s-ai-challenge.github.io/}), we aim to predict the biweekly average statistics of weeks 3-4 and weeks 5-6, which correspond to lead times of $2$ weeks and $4$ weeks, respectively. We construct ERA5-S2S, a new dataset from $5.625\degree$ ERA5 that has the same input variables, but the output variables are averaged from the lead time to $2$ weeks ahead into the future.  

We compare ClimaX with ResNet, UNet, and Cli-ViT on the S2S prediction of four target variables: T850, T2m, U10, and V10. 
\Cref{tab:s2s_prediction} compares the RMSE of ClimaX and the baselines. ClimaX achieves the lowest error for all variables, and the performance gap with the best baseline UNet is larger at increasing lead times. ClimaX also has significant performance gains over its scratch-trained counterpart Cli-ViT, showing the effectiveness of our pretraining procedure in capturing features that are generally useful for various temporal prediction tasks.

\begin{table}[t]
\centering
\caption{RMSE of ClimaX and baselines on $5.625\degree$ ERA5-S2S prediction tasks.
}
\label{tab:s2s_prediction}
\resizebox{0.8\columnwidth}{!}{
\begin{tabular}{@{}lcccccccc@{}}
\toprule
               & \multicolumn{2}{c}{T850}                                      & \multicolumn{2}{c}{T2m}                                       & \multicolumn{2}{c}{U10}                                       & \multicolumn{2}{c}{V10}                                       \\ \cmidrule(l){2-9} 
               & \multicolumn{1}{c}{Weeks 3-4} & \multicolumn{1}{c}{Weeks 5-6} & \multicolumn{1}{c}{Weeks 3-4} & \multicolumn{1}{c}{Weeks 5-6} & \multicolumn{1}{c}{Weeks 3-4} & \multicolumn{1}{c}{Weeks 5-6} & \multicolumn{1}{c}{Weeks 3-4} & \multicolumn{1}{c}{Weeks 5-6} \\ \midrule
Resnet         & 2.12                          & 2.13                          & 1.88                          & 2.16                          &           1.91                    &             1.94                  &           1.52                    &             1.59                  \\
Unet           & 1.91                          &      1.95                         & 1.67                          & 1.79                          &               1.85                &                1.90               &           1.52                    &               1.57                \\
Cli-ViT & 1.96                          & 1.96                          & 1.79                          & 1.90                          &                  1.83             &               1.92                &            1.51                   &                 1.56              \\
ClimaX         & \textbf{1.89}                          & \textbf{1.92}                          & \textbf{1.66}                          & \textbf{1.70}                          &               \textbf{1.81}                &             \textbf{1.86}                  &                \textbf{1.50}               &                 \textbf{1.54}              \\ \bottomrule
\end{tabular}
}
\end{table}

\subsection{Climate projection}
To further test the generality of ClimaX, we evaluate the model on ClimateBench~\citep{watson2022climatebench}, a recent benchmark designed for testing machine learning models for climate projections. The goal of ClimateBench is to predict the annual mean global distributions of surface temperature, diurnal temperature range, precipitation, and the $90$th percentile of precipitation, given the four anthropogenic forcing factors: carbon dioxide (CO$_2$), sulfur dioxide (SO$_2$), black carbon (BC), and methane (CH$_4$). We note that this is not a temporal modeling task, as we do not predict the future given the past. Instead, we answer questions like \emph{what will be the annual mean temperature for a specified CO$_2$ level?} In particular, note that the input variables and the task itself are completely different from pretraining.

\begin{figure}[t]
    \centering
    \includegraphics[width=0.95\textwidth]{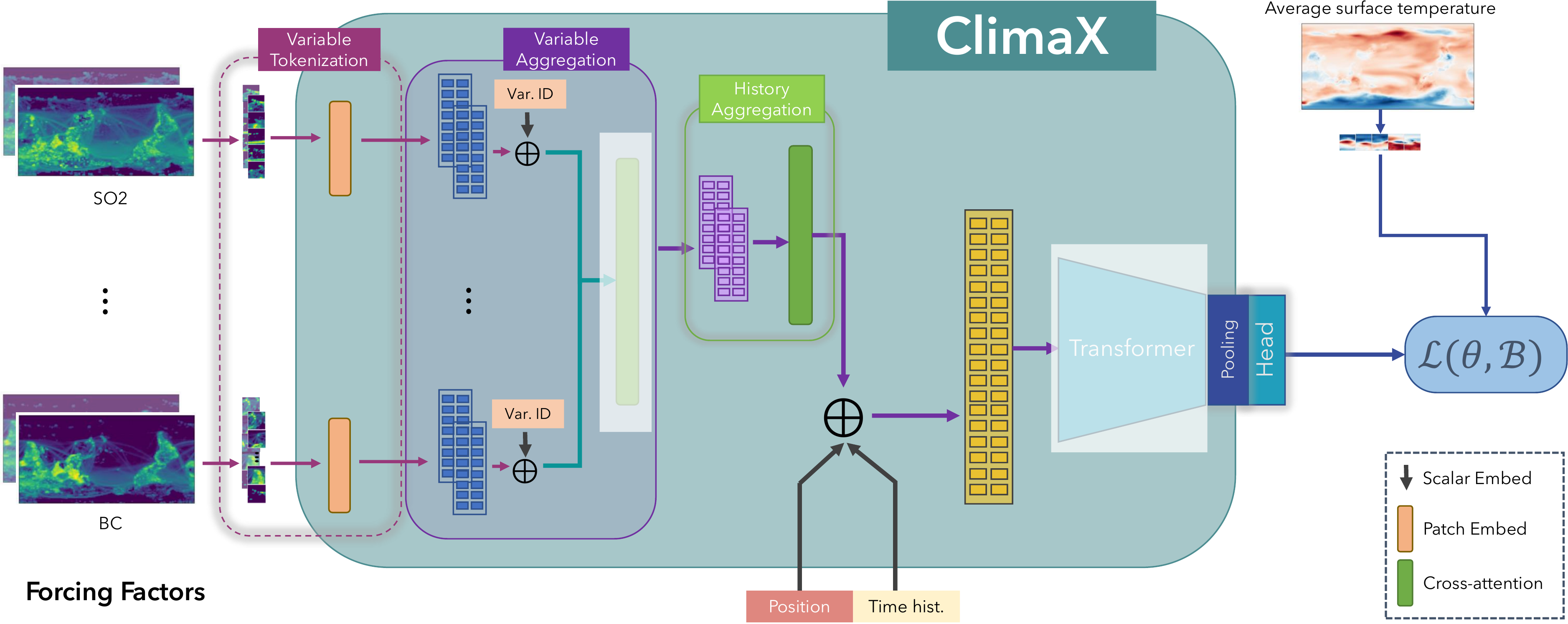}
    \caption{Finetuning pipeline for ClimateBench. A different set of input and output variables requires different embedding layers and prediction heads. Attention layers can be frozen or finetuned.}
    \label{fig:climatebench-Finetuning}
\end{figure}

\Cref{fig:climatebench-Finetuning} illustrates the finetuning pipeline of ClimaX for ClimateBench. As the input and output variables are unseen during pretraining, we replace the pretrained embedding layers and prediction heads with newly initialized networks, while keeping the attention layers and the variable aggregation module. We consider two finetuning protocols, in which we either freeze\footnote{We finetune the LayerNorm in ClimaX$_{\text{frozen}}$, as suggested by~\citet{lu2021pretrained}.} (ClimaX$_{\text{frozen}}$) or finetune (ClimaX) the attention layers. In addition, we introduce two components to the pipeline in \Cref{fig:pretrain}. We use a history of the preceding ten years of the forcing factors to make predictions for a particular year, creating an input of shape $T \times V \times H \times W$. 
Each time slice of the input goes through variable tokenization, variable aggregation, and the attention layers as usual, which output a feature tensor of shape $T \times h \times w \times D$, where $D$ is the embedding size. The feature tensor then goes through a global average pooling layer, reducing the dimension to $T \times D$. Finally, the $10$-year history is aggregated using a cross-attention layer before being fed to the prediction head, which linearly transforms the $D$-dimensional feature vector to a $H \times W$ map. The history aggregation and the global pooling modules are the two additions to the original ClimaX architecture. These architectural designs are inspired by the neural network baseline in~\citep{watson2022climatebench}.

We compare ClimaX with ClimaX$_\text{frozen}$, Cli-ViT, and the best baseline from ClimateBench. Following~\citep{watson2022climatebench}, we use the standard mean squared error (\Cref{eq:lat_mse} without the weighting term) as the loss function. We evaluate all methods on RMSE, NRMSE$_s$ (Spatial), NRMSE$_g$ (Global), and Total = NRMSE$_s$ + 5 $\times$ NRMSE$_g$~\citep{watson2022climatebench}. Details of the metrics are in \Cref{sec:app:metrics}. \Cref{tab:climate_bench} shows the results. ClimaX$_\text{frozen}$ performs the best in predicting two temperature-related variables, followed by ClimaX. This shows that the pretrained attention layers can serve as a strong feature extractor in seemingly unrelated tasks. Where downstream data is scarce (ClimateBench has only $754$ data points), further finetuning the attention layer can lead to overfitting and thus slightly hurt the performance. In two precipitation-related tasks, ClimaX$_\text{frozen}$ slightly underperforms ClimateBench baseline in terms of NRMSE$_s$ and NRMSE$_g$ but outperforms on RMSE. We hypothesize that this was because ClimaX did not observe the precipitation variable during pretraining, which has very different behaviors from other variables.

\begin{table}[t]
\centering
\caption{Performance of ClimaX and the baselines on ClimateBench. Spatial and Global denote the normalized root mean squared error NRMSE$_s$ and the NRMSE of the global mean NRMSE$_g$, respectively. Total is a weighted combination of Spatial and Global.
}
\label{tab:climate_bench}
\resizebox{\columnwidth}{!}{
\begin{tabular}{@{}lcccccccccccccccc@{}}
\toprule
                         & \multicolumn{4}{c}{Surface temperature} & \multicolumn{4}{c}{Diurnal temperature range} & \multicolumn{4}{c}{Precipitation} & \multicolumn{4}{c}{90th percentile precipitation} \\ \cmidrule(lr){2-5} \cmidrule(lr){6-9} \cmidrule(lr){10-13} \cmidrule(lr){14-17} 
                         & Spatial   & Global   & Total   & RMSE   & Spatial     & Global    & Total     & RMSE    & Spatial  & Global  & Total & RMSE & Spatial      & Global      & Total     & RMSE     \\ \midrule
ClimateBench-NN (reproduced)    & 0.123     & 0.080    & 0.524   & 0.404    & 7.465       & 1.233     & 13.632    & 0.150     & 2.349    &  \textbf{0.151}   & \textbf{3.104} & 0.553  & 3.108        & \textbf{0.282}       & 4.517     & 1.594      \\
ClimateBench-NN (paper)   & 0.107     & 0.044    & 0.327   & N/A    & 9.917       & 1.372     & 16.778    & N/A     & \textbf{2.128}    & 0.209   & 3.175 & N/A  & \textbf{2.610}        & 0.346       &  \textbf{4.339}     & N/A      \\
Cli-ViT                  & 0.086     & 0.044    & 0.305   & 0.362  &    6.997         &     1.759      &      15.792     &    0.146     &      2.224    &    0.241     &     3.430  &   0.550   &        2.800      &      0.329       &     4.447      &     1.579     \\
ClimaX         & 0.086     &  \textbf{0.043}    & 0.300   & 0.362  &      7.148       &      0.961     &     11.952      &    0.147     &     2.360     &    0.206     &    3.390   &   0.554   &      2.739        &       0.332      &     4.397      &      1.575    \\ 
ClimaX$_\text{frozen}$ & \textbf{0.085}     &  \textbf{0.043}    &  \textbf{0.297}   & \textbf{0.360}  &       \textbf{6.688}      &     \textbf{0.810}      &     \textbf{10.739}      &   \textbf{0.144}      &     2.193     &    0.183     &     3.110  &   \textbf{0.549}   &        2.681      &      0.342       &     4.389      &      \textbf{1.572}    \\
\bottomrule
\end{tabular}
}
\end{table}

\subsection{Climate model downscaling}
Climate models are often run at coarse grids due to their high computational cost. Although these predictions are useful in understanding large-scale climate trends, they do not provide sufficient detail to analyze regional and local phenomena. Downscaling aims to obtain higher-resolution projections and reduce biases from the outputs of these models. 
To evaluate the applicability of ClimaX to the task of climate model downscaling, we construct a new dataset based on CMIP6 and ERA5 data sources for coarse inputs and higher resolution targets.
Specifically, we use all MPI-ESM, a dataset from CMIP6, and its variables listed in \Cref{tab:cmip6_data} at $5.625\degree$ as input, and train separate models to downscale to each ERA5 target variable at $1.40625\degree$. We compare ClimaX with Cli-ViT and the two CNN baselines, UNet and ResNet, as most recent deep downscaling methods~\citep{vandal2017deepsd,rodrigues2018deepdownscale,hohlein2020comparative,vandal2019intercomparison,liu2020climate} are based on convolution. We were not able to compare with YNet~\citep{liu2020climate}, the current best method on deep downscaling as we did not have access to high-resolution auxiliary data such as elevation and topographical information.
For all methods, we first bilinearly interpolate the input to match the resolution of the desired output before feeding it to the model. We evaluate all methods on RMSE, Pearson correlation, and Mean bias, which were commonly used in existing deep downscaling works~\citep{vandal2017deepsd,liu2020climate}.
Details of the metrics are in \Cref{sec:app:metrics}. 

\begin{figure}[t]
    \centering
    \includegraphics[width=\textwidth]{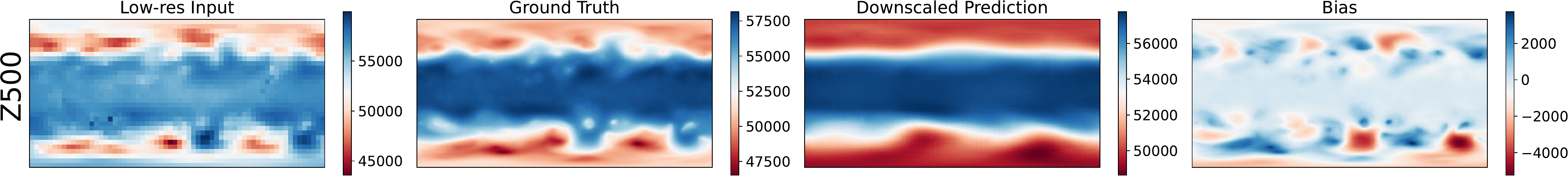}
    \includegraphics[width=\textwidth]{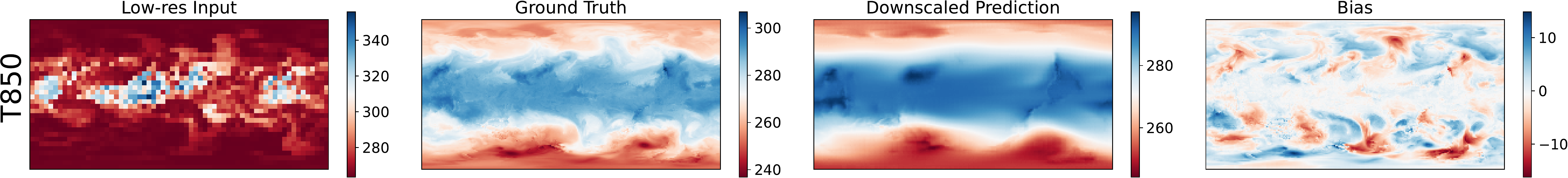}
    \includegraphics[width=\textwidth]{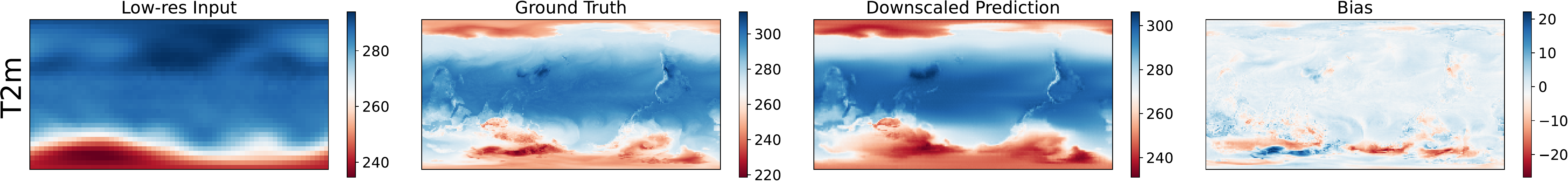}
    \caption{Example visualizations of downscaled prediction of key variables by ClimaX.}
    \label{fig:downscaleviz}
\end{figure}

\Cref{tab:downscaling} compares ClimaX and the baselines quantitatively. ClimaX achieves the lowest RMSE and a mean bias closest to $0$ for all three target variables, and performs similarly to the baselines in terms of Pearson correlation.
While pretrained to perform forecasting, ClimaX has successfully captured the spatial structure of weather data, which helps in downstream tasks like downscaling.
\Cref{fig:downscaleviz} visualizes the downscaled predictions of ClimaX for the three target variables. The input is at a much lower resolution and contains a lot of bias compared to the ground truth. While the prediction is missing some fine details, it has successfully captured the general structure of the ERA5 data and removed input biases.

\begin{table}[t]
\centering
\caption{Performance of ClimaX and the baselines on downscaling from MPI-ESM ($5.625\degree$) to ERA5 ($1.40625\degree$).
}
\label{tab:downscaling}
\sisetup{detect-weight=true,detect-inline-weight=math}
\resizebox{\columnwidth}{!}{
\begin{tabular}{@{}lSSSSSSSSSSSSSSS@{}}
\toprule
               & \multicolumn{3}{c}{Z500}     & \multicolumn{3}{c}{T850}   & \multicolumn{3}{c}{T2m}    & \multicolumn{3}{c}{U10}    & \multicolumn{3}{c}{V10}    \\ \cmidrule(lr){2-4} \cmidrule(lr){5-7} \cmidrule(lr){8-10} \cmidrule(lr){11-13} \cmidrule(lr){14-16} 
               & {RMSE}   & {Pearson} & {Mean bias} & {RMSE} & {Pearson} & {Mean bias} & {RMSE} & {Pearson} & {Mean bias} & {RMSE} & {Pearson} & {Mean bias} & {RMSE} & {Pearson} & {Mean bias} \\ \midrule
ResNet         & 825.75 & 0.96    & -108.54   & 3.60 & 0.96    & 0.19      & 2.89 & 0.98    & 0.14      & 4.05 & 0.65    & 0.06      & 4.11 & 0.45    & 0.09      \\
UNet           & 858.35 & 0.95    & 35.10     & 3.66 & 0.96    & -0.34     & 2.95 & 0.98    & 0.16      & 4.09 & 0.64    & -0.06     & 4.13 & 0.44    & 0.08      \\
Cli-ViT & 811.61 & 0.96    & -54.32    & 3.58 & 0.97    & -0.29     & 2.80 & 0.99    & -0.06     & 4.01 & 0.66    & -0.08     & 4.07 & 0.47    & 0.01      \\
ClimaX         & \bfseries 807.43 & 0.96    & \bfseries{2.70}      & \bfseries{3.49} & 0.97    & \bfseries -0.11     &  \bfseries 2.79 & 0.99    &  \bfseries -0.06     &  \bfseries 3.99 &  \bfseries 0.66    &  \bfseries 0.04      &  \bfseries 4.06 & 0.47    & -0.02     \\ \bottomrule
\end{tabular}
}
\end{table}

\subsection{Scaling laws analysis} \label{sec:scaling}
\begin{figure}[t]
    \centering
    \includegraphics{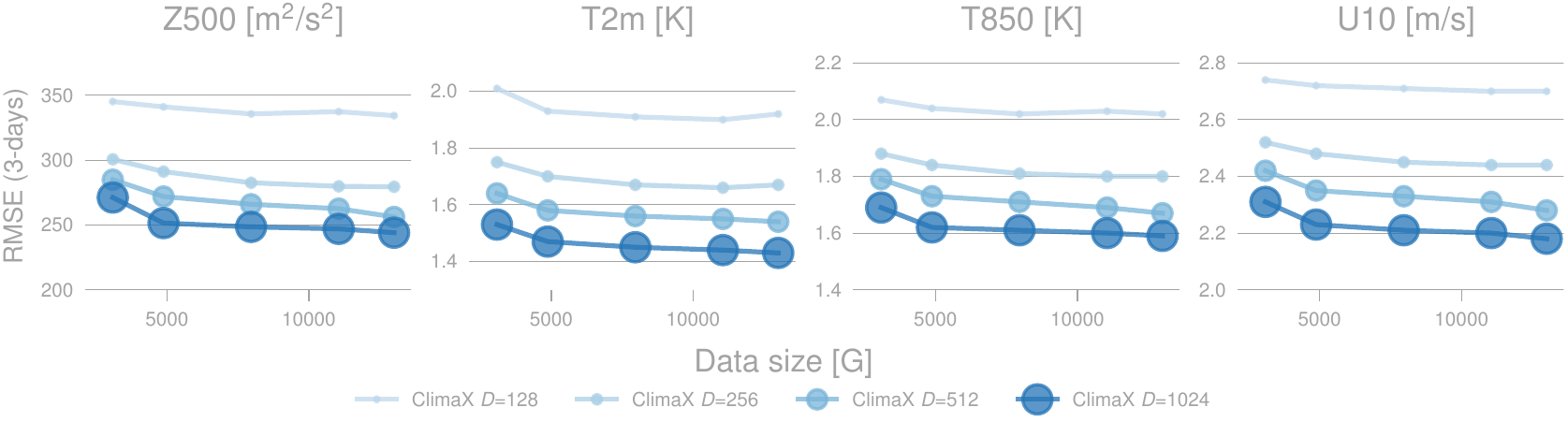}
    \caption{Error on ERA5 3-day forecasting for different variables with respect to CMIP6 5.625$\degree$ data seen during pre-training. Bigger models are more sample efficient.}
    \label{fig:scaling}
\end{figure}

Transformers have shown favorable scaling properties for language \citep{kaplan2020scaling,hoffmann2022training}, vision \citep{zhai2022scaling}, or even multi-modal tasks~\citep{henighan2020scaling,hendricks2021decoupling,reed2022gato}. That is, their performance improves with respect to data size and model capacity given sufficient compute. 
In this section, we study the scaling laws of ClimaX in weather forecasting. \Cref{fig:scaling} presents the performance of ClimaX as a function of data size and model capacity. The $x$-axis is the pretraining data size measured in Gigabytes, which corresponds to $1$ to $5$ CMIP6 datasets, and the $y$-axis shows the RMSE of ClimaX on the 3-day forecasting task. We compare four ClimaX models with different capacities by varying the embedding dimension from $128$ to $1024$. All experiments are conducted on the $5.625\degree$ data. The error rate of the two biggest models decreases consistently as we increase the data and model size. This highlights the unique ability of ClimaX in learning from diverse and heterogeneous data sources, which allows us to further improve the performance by simply pretraining on more data. However, the two smaller models do not scale as well as the bigger ones, where increasing data size does not gain much improvement or can sometimes hurt performance. This result shows that larger models not only perform better but are also more data efficient.

\begin{figure}[t]
    \centering
    \begin{subfigure}[b]{0.49\textwidth}
        \includegraphics[width=\textwidth]{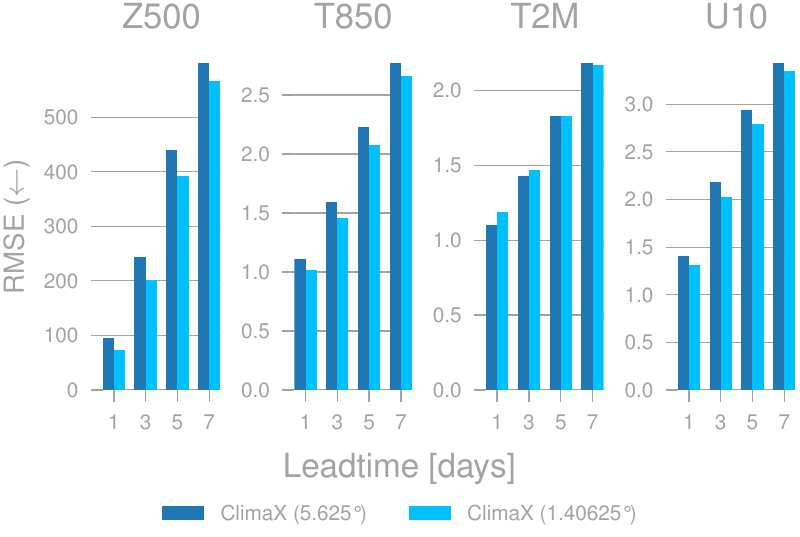}
    \end{subfigure}
    \begin{subfigure}[b]{0.49\textwidth}
        \includegraphics[width=\textwidth]{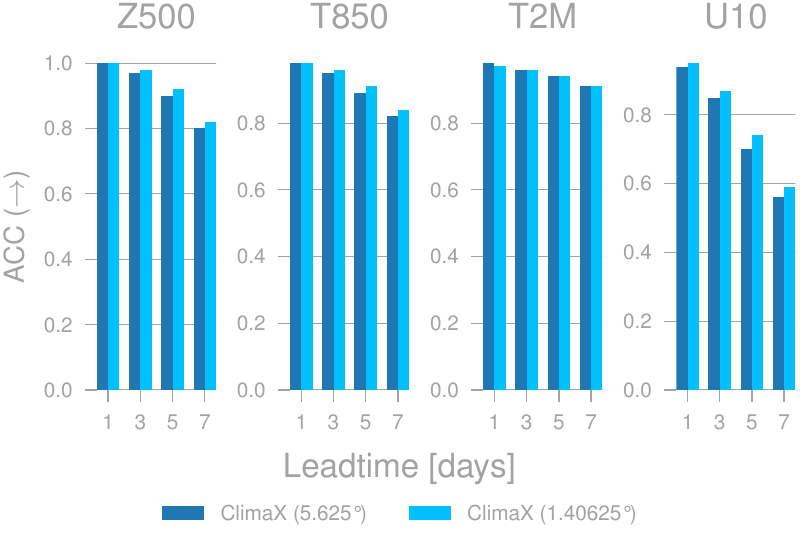}
    \end{subfigure}
    \caption{Scaling performance with respect to data resolution. Despite a larger patch size, ClimaX ($1.40625\degree$) achieves consistently better performance than the low-resolution model on almost all tasks, except for T2m forecast at 1 day and 3 days lead times.
    }
    \label{fig:scaling-res}
\end{figure}

In addition to data size and model capacity, data resolution is another important scaling dimension in the context of weather and climate. In many vision tasks such as classification, understanding the general, high-level structure of the image is sufficient to make accurate predictions. To model the underlying complex physical processes that govern weather and climate, however, it is important for a model to look at fine-grained details of the input in order to understand the spatial and temporal structure of data as well as the interactions between different variables. High-resolution data contains finer details and local processes of weather conditions that are not present in the low-resolution data, and thus provides stronger signals for training deep learning models. \Cref{fig:scaling-res} compares the performance of ClimaX pretrained and finetuned on $5.625\degree$ and $1.40625\degree$ data on global forecasting. Except for T2m at $1$ day and $3$ days lead times, ClimaX ($1.40625\degree$) consistently achieves lower RMSE and higher ACC than the low-resolution model. We note that for the high-resolution data we have to use a larger patch size ($4$ compared to $2$ for low-resolution data) due to lack of memory issue. We can further improve the performance of ClimaX on the $1.40625\degree$ data by reducing the patch size, as the model is able to capture better details.

\subsection{Ablation studies} \label{sec:ablation}

In the main forecasting results, we finetune a separate ClimaX model for each target variable at each lead time, as we found this protocol led to the best performance. However, this can be computationally expensive, as finetuning cost scales linearly with respect to the number of target variables and lead times. In this section, we consider different finetuning alternatives to investigate the trade-off between computation and performance.

\begin{figure}[t]
    \centering
    \includegraphics[width=\textwidth]{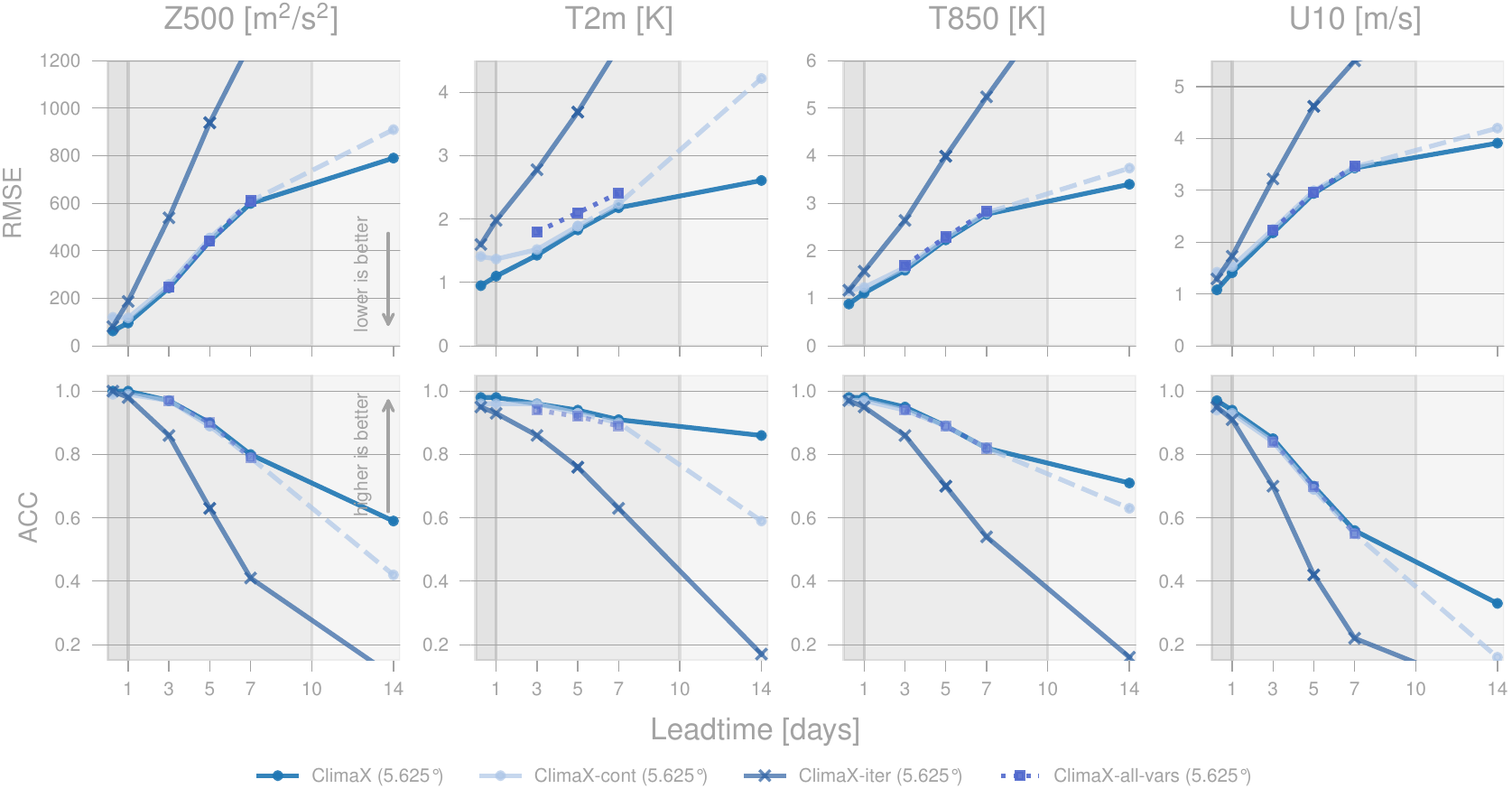}
    \caption{Performance of ClimaX and its variations on weather forecasting. ClimaX-cont is a lead-time-conditioned model that we finetune to make predictions at 6 hours to 7 days. ClimaX-iter forecasts at a $6$-hour lead time and rolls out the predictions to forecast at longer horizons. ClimaX-all-vars predicts the future conditions of all variables in the input at particular lead-times.}
    \label{fig:ablation-alternative-finetuning}
\end{figure}

\subsubsection{Should we finetune ClimaX for each variable separately or all at once?}
Instead of finetuning ClimaX for each target variable separately, we could alternatively finetune once to predict all variables in the input simultaneously, which we denote as ClimaX-all-vars. \Cref{fig:ablation-alternative-finetuning} shows that ClimaX-all-vars achieves comparable performance to ClimaX in most of the tasks and only underperforms for forecasting T2m. This suggests that with a limited budget, one can finetune ClimaX to predict all target variables at the same time without losing much performance.

\subsubsection{Should we do iterative forecast or direct forecast?}
To avoid finetuning a different model for each lead time, we can finetune ClimaX to make predictions at a short horizon such as $6$ hours, and roll out the predictions during inference to make forecasts at longer horizons. We call this model ClimaX-iter, where \emph{iter} stands for iterative prediction~\citep{rasp2020weatherbench}. We note that in order to roll out more than one step, ClimaX-iter must predict for all input variables, or in other words. This provides the benefit of finetuning a single model that can predict for any target variable at any lead time. \Cref{fig:ablation-alternative-finetuning} shows that ClimaX-iter works reasonably well up to 1-day prediction, but the performance degrades significantly at longer lead times. This is not surprising, because ClimaX-iter is not finetuned to predict multiple steps into the future, leading to quick error accumulation. One can employ a multi-step objective for finetuning as in~\citet{pathak2022fourcastnet} to achieve better results.

\subsubsection{Can we finetune ClimaX to work for all lead times?}
Another way to avoid finetuning for each lead time separately is to finetune a lead-time-conditioned model. Specifically, during finetuning, we randomize the lead time from $6$ hours to $7$ days, resembling the pretraining setting. Note that unlike ClimaX-iter, we still have to finetune a separate model for each target variable. We call this model ClimaX-cont, wherein \emph{cont} stands for \emph{continuous}, a standard term used in previous works~\citep{rasp2020weatherbench}. \Cref{fig:ablation-alternative-finetuning} shows that ClimaX-cont performs competitively on $6$-hour to $7$-day forecasting, but fails to extrapolate to $2$ weeks and $1$ month lead times that are unseen during training. 
One can also randomize the lead time from $6$ hours to $1$ month, but that means the model sees much fewer data points for each target lead time, potentially hurting the performance.

The cost for finetuning each set of weights is a constant $C$, which is about $15$ hours on an $8 \times \text{V}100s$. Among different finetuning protocols, ClimaX is the most expensive, whose total cost is $C \times \#variables \times \#lead\_times$, scaling linearly with the number of target variables and lead times. Following ClimaX are ClimaX-all-vars and ClimaX-cont, whose total costs are $C \times \#lead\_times$ and $C \times \#variables$, respectively. Finally, ClimaX-iter is the cheapest finetuning protocol, where we only have to finetune a single model that works for all target variables and at all lead times. The performance is proportional to the computational cost, as ClimaX is the best performing model, while ClimaX-iter is the worst.

\section{Discussion and Future Work}\label{sec:discuss}

The scaling of datasets, model architectures, and computation has resulted in a transformative impact in various subdisciplines of artificial intelligence, from natural language and speech processing to computer vision, as well as scientific applications in biology and chemistry. In particular, it has led to the emergence of general-purpose foundation models that are trained on large datasets and compute clusters, and can be easily adapted to a variety of downstream tasks efficiently, both in terms of compute and data supervision.
Our work represents a pioneering effort to enable such broad scaling and generality in data-driven models for weather and climate. This approach goes beyond the limitations of both traditional numerical modeling and existing data-driven forecasting methods. Unlike ClimaX, numerical models scale only in terms of computation and not in terms of dataset size, whereas existing data-driven models are typically limited to specific tasks and lack general-purpose applicability across a wide range of tasks.

In addition to traditional considerations in language and vision, foundation models like ClimaX open up new opportunities for scaling through the use of simulation datasets and grid resolutions. To simplify our approach, we chose to use pretraining datasets that include standard variables that have been benchmarked in previous research on data-driven forecasting~\cite{rasp2020weatherbench,pathak2022fourcastnet}. Additionally, we avoided datasets that simulate future scenarios under different forcings to prevent any potential leakage for the climate projection task. Future research could explore incorporating both observational and simulated datasets that include a wider range of climate variables, higher spatiotemporal resolutions, and even extend into future scenarios.
Further, we showed that resolution plays a crucial role in scaling of ClimaX. Due to our compute restrictions, we trained ClimaX on low to moderate resolutions. 
Nevertheless, our empirical trends suggest that scaling to higher resolutions ($0.25\degree$) is likely to lead to even better results.

Scaling efforts in the future can benefit from better sequence modeling architectures, especially those designed for multimodal spatiotemporal inputs.
As we saw in ClimaX, the number of channels for climate datasets is much greater than those handled for standard multimodal settings (e.g., audio-video, vision-language models). 
Moreover, in practice, there is also a significant range of resolutions across different climate datasets.
This heterogeneity drastically increases the raw length of input sequences for standard architectures such as ViT.
In the future, we believe that investigating single multi-scale architectures (e.g., \citep{fan2021multiscale}) can potentially aid in scaling to such diverse multi-resolution and multi-modal datasets by learning to infer features relevant to atmospheric phenomena at increasing spatial resolutions.

In conclusion, we believe that the generality of our approach has potential applications beyond the tasks considered in this work. It would be interesting to explore the generalization of a pretrained ClimaX backbone to other Earth systems science tasks, such as predicting extreme weather events~\citep{miralles2019land, sillmann2017understanding} and assessing anthropogenic contributions to climate change~\citep{rosenzweig2008attributing,hook2013depletion}, as well as broader domains that are closely tied to weather and climate conditions, such as agriculture, demography, and actuarial sciences.

\begin{acks}
    We would like to thank ECMWF for enabling this line of research with accessible public datasets, contributors of WebPlotDigitizer~\citep{Rohatgi2022} for making it easier to build \Cref{tab:rmse-compare,tab:acc-compare}, and numerous other open-source libraries, notably numpy~\citep{harris2020array} and PyTorch~\citep{paszke2019pytorch}.
    Some icons in \cref{fig:climax-feature} by Freepik, smalllikeart, and GOWI from \url{flaticon.com}.
\end{acks}

\printbibliography

\appendix

\clearpage

\section{Model}\label{sec:app:arch}
This section presents the implementation details and hyperparameters of ClimaX and the two CNN baselines UNet and ResNet.

\subsection{ClimaX}
\subsubsection{Implementation details}
ClimaX receives a tensor of shape $V \times H \times W$ and outputs a tensor of shape $V' \times H \times W$, where the number of input and output variables $V$ and $V'$ can vary between different datasets\footnote{The spatial resolution $H \times W$ can also vary. In that case, we employ the common practice of interpolating the positional embedding, and everything else remains the same~\citep{dosovitskiy2020image,touvron2021training}.}. To do that, we assume a set $\mathcal{V}$ that contains all possible variables we could encounter during pretraining and finetuning. Each variable in $\mathcal{V}$ has a separate token embedding layer.

The variable tokenization module tokenizes the input to a sequence of $V \times h \times w$ tokens, with each token being a vector of size $p^2$. After that, for each token, we extract the corresponding embedding layer that transforms the token to a vector of dimension $D$. Each embedding layer is a single convolution layer with $in\_channels=1, out\_channels=D, kernel\_size=p, stride=p$. This results in a tensor of shape $V \times h \times w \times D$.

To differentiate between tokens of different input variables, we add the sequence with a \emph{variable positional embedding}, which is a tensor of shape $|\mathcal{V}| \times D$. For each input variable, we extract the corresponding variable positional embedding to add to its tokens. After that, all tokens go through the variable aggregation module, which outputs a tensor of shape $h \times w \times D$.

The tokens are then fed to the attention layers, which output a tensor of the same shape $h \times w \times D$. The prediction head takes each token of dimension $D$ and maps it to a vector of dimension $|\mathcal{V}| \times p^2$, and the output is reshaped to $|\mathcal{V}| \times H \times W$. Finally, we extract predictions of $V'$ target variables and compute the loss.

\subsubsection{Hyperparameters}

\begin{table}[h]
\centering
\caption{Default hyperparameters of ClimaX}
\begin{tabular}{@{}lll@{}}
\toprule
Hyperparameter   & Meaning                                                                                                  & Value                                                                                      \\ \midrule
$\mathcal{V}$    & Default variables                                                                                        & All ERA5 variables in Table~\ref{tab:era5_data}                                                                                           \\
$|\mathcal{V}|$  & Number of default variables                                                                              & 48                                                                                         \\
$p$              & Patch size                                                                                               & \begin{tabular}[c]{@{}l@{}}$2$ for  $5.625\degree$\\ $4$ for $1.40625\degree$\end{tabular} \\
$D$              & Embedding dimension                                                                                      & $1024$                                                                                     \\
Depth            & Number of ViT blocks                                                                                     & $8$                                                                                        \\
\# heads         & Number of attention heads                                                                                & $16$                                                                                       \\
MLP ratio        & \begin{tabular}[c]{@{}l@{}}Determine the hidden dimension of\\ the MLP layer in a ViT block\end{tabular} & $4$                                                                                        \\
Prediction depth & Number of layers of the prediction head                                                                  & $2$                                                                                        \\
Hidden dimension & Hidden dimension of the prediction head                                                                  & $1024$                                                                                     \\
Drop path        & For stochastic depth~\citep{huang2016deep}                                                               & $0.1$                                                                                      \\
Dropout          & Dropout rate                                                                                             & $0.1$                                                                                      \\ \bottomrule
\end{tabular}
\end{table}

\subsection{CNN Baselines} \label{sec:app:cnn_arc}

\subsubsection{ResNet Hyperparameters}
We use the following hyperparameters for ResNet in all of our experiments.
\begin{table}[h!]
\centering
\caption{Default hyperparameters of ResNet}
\begin{tabular}{@{}lll@{}}
\toprule
Hyperparameter   & Meaning                                          & Value \\ \midrule
Padding size     & Padding size of each convolution layer           & $1$   \\
Kernel size      & Kernel size of each convolution layer            & $3$   \\
Stride           & Stride of each convolution layer                 & $1$   \\
Hidden dimension & Number of output channels of each residual block & $128$ \\
Residual blocks  & Number of residual blocks                        & $28$  \\
Dropout          & Dropout rate                                     & $0.1$ \\ \bottomrule
\end{tabular}
\end{table}

\subsubsection{UNet Hyperparameters}
We borrow our UNet implementation from \href{https://github.com/microsoft/pdearena/blob/main/pdearena/modules/twod_unet.py}{PDEArena}~\citep{gupta2022towards}. 
We use the following hyperparameters for UNet in all of our experiments.
\begin{table}[h!]
\centering
\caption{Default hyperparameters of UNet}
\begin{tabular}{@{}lll@{}}
\toprule
Hyperparameter          & Meaning                                                                                                  & Value          \\ \midrule
Padding size            & Padding size of each convolution layer                                                                   & $1$            \\
Kernel size             & Kernel size of each convolution layer                                                                    & $3$            \\
Stride                  & Stride of each convolution layer                                                                         & $1$            \\
Channel multiplications & \begin{tabular}[c]{@{}l@{}}Determine the number of output channels\\ for Down and Up blocks\end{tabular} & $[1, 2, 2, 4]$ \\
Blocks                  & Number of blocks                                                                                         & $2$            \\
Use attention           & If use attention in Down and Up blocks                                                                   & False          \\
Dropout                 & Dropout rate                                                                                             & $0.1$          \\ \bottomrule
\end{tabular}
\end{table}

\subsubsection{Other implementation details}
Following the implementation of ResNet in~\citet{rasp2020weatherbench,rasp2021data,ernst2021structured}, we found the following details important for the performance of both CNN baselines:
\begin{itemize}
    \item Use Batch normalization
    \item Use Leakyrelu with a slope of $0.3$ as the activation function
    \item Postnorm instead of Prenorm
    \item Use periodic convolutions in the longitude direction but not the latitude direction. 
    \item Use a kernel size of $7$ in the first CNN layer.
\end{itemize}

\section{Training details}

\paragraph{Data normalization}
We normalized all inputs during pre-training as well as fine-tuning. 
For each variable, at each pressure level (for atmospheric variables), we compute the mean and standard deviation to normalize them to zero mean and unit variance. We de-normalize the predictions to get back to the original range before computing evaluation metrics.

\paragraph{Software and hardware stack}
We use PyTorch \citep{pytorch2019}, \texttt{timm}~\citep{rw2019timm}, \texttt{numpy}~\citep{harris2020array} and \texttt{xarray} \citep{Hoyer_2017} to manage our data and model training.
We used 32GB NVIDIA V100 devices for training. For pretraining we distribute the batch across 80 V100s on AzureML. We leverage \texttt{fp16} floating point precision in our model.  

\subsection{Pretraining}

\subsubsection{Objective}
We use the loss function in \Cref{eq:lat_mse} for pretraining.

\subsubsection{Optimization}
We used the AdamW optimizer~\citep{kingma2014adam,loshchilov2017decoupled} with parameters ($\beta_1 = 0.9, \beta_2 = 0.95$). We used weight decay of $1e-5$ for all parameters except for the positional embedding. We used a learning rate of $5e-4$, with a linear warmup schedule for $10000$ steps ($5$ epochs), followed by a cosine-annealing schedule for $190000$ steps ($95$ epochs).

\subsection{Finetuning}
\subsubsection{Objective}
We use lat-weighted MSE in \Cref{eq:lat_mse} for finetuning ClimaX in temporal forecasting and downscaling tasks. In ClimateBench, we finetune using standard MSE without the weighting term, as this led to better results and was suggested by~\citep{watson2022climatebench}.

\subsubsection{Optimization}
For all tasks, we used AdamW with parameters ($\beta_1 = 0.9, \beta_2 = 0.999$). We used weight decay of $1e-5$ for all parameters except for the positional embedding. We used a linear warmup schedule for $10000$ steps ($5$ epochs), followed by a cosine-annealing schedule for $90000$ steps ($45$ epochs). The learning rate for each task is as follows:
\begin{table}[h]
\centering
\caption{Learning rate for finetuning ClimaX in different downstream tasks}
\begin{tabular}{@{}ll@{}}
\toprule
Task                & Learning rate \\ \midrule
Weather forecasting & $5e-7$        \\
Climate projection  & $5e-4$        \\
Climate downscaling & $5e-5$        \\ \bottomrule
\end{tabular}
\end{table}

We used a small learning rate for weather forecasting as the task resembles pretraining. For downscaling, we used a larger learning rate, as the nature of the task is different from pretraining, even though the input variables are similar. In climate projection, we needed to initialize new weights for the embedding layers and prediction heads, and thus used a similar learning rate to training from scratch.

\section{Datasets}\label{sec:app:data}

\subsection{CMIP6-ClimaX} \label{sec:app:cmip6_details}
We created CMIP6-ClimaX for pretraining ClimaX, which consists of $5$ datasets from the CMIP6 project. We downloaded the datasets from the official CMIP6 search interface at \href{https://esgf-data.dkrz.de/search/cmip6-dkrz/}{https://esgf-data.dkrz.de/search/cmip6-dkrz/}. These datasets share the following attributes:
\begin{itemize}
    \item Experiment ID: historical
    \item Table ID: 6hrPlevPt, i.e., 6-hourly data on pressure levels.
    \item Variant label: r1i1p1f1. The variant label distinguishes among closely related simulations by a single model, in which ``r'' specifies the initial condition, ``i'' specifies the observational dataset and initialization method used for determining the initial condition, ``p'' specifies the perturbed physics version of the model, and ``f'' specifies the forcing index.
\end{itemize}
All datasets have a temporal coverage from $1850$ to $2015$ and a temporal resolution of $6$ hours. We chose these datasets as they contain similar climate variables at similar vertical levels to ERA5.
We note that there are more than $5$ datasets from CMIP6 that suit our selection criteria, but we were not able to download others due to some issues on the data servers. We regridded these datasets to $5.625\degree$ and $1.40625\degree$ using the xesmf Python package~\citep{zhuang2018xesmf} using bilinear interpolation.
We provide a detailed description of these $5$ data sources and the available variables we used to construct CMIP6-ClimaX in \Cref{tab:cmip6_data}. 

\begin{table}[h!]
\centering
\caption{Resolution and variables of CMIP6-ClimaX dataset used for pretraining. \emph{Static} represents variables don't depend on time, \emph{Single}  represents surface variables, and \emph{Atmospheric} represents time-varying atmospheric properties at the chosen altitudes.}
\label{tab:cmip6_data}
\begin{tabular}{@{}lllll@{}}
\toprule
\multirow{2}{*}{Data Source} & \multirow{2}{*}{Original resolution} & \multicolumn{3}{c}{Variables}                                                                                                       \\ \cmidrule{3-5}
{} & {} & {Type} & {Abbrev.} & {Levels} \\\midrule
\multirow{8}{*}{MPI}     & \multirow{8}{*}{100km}               &  Single & t2m \\ 
        & & Single &  u10 \\ 
        & & Single & v10\\ 
        & & Atmospheric & z & 50, 250, 500, 600, 700, 850, 925\\ 
        & & Atmospheric & u & 50, 250, 500, 600, 700, 850, 925\\ 
        & & Atmospheric & v & 50, 250, 500, 600, 700, 850, 925\\
        & & Atmospheric & t & 50, 250, 500, 600, 700, 850, 925\\ 
        & & Atmospheric & q & 50, 250, 500, 600, 700, 850, 925\\ \midrule
\multirow{6}{*}{Tai}     & \multirow{6}{*}{100km}               & Single & t2m\\ 
        & & Atmospheric & z & 250, 500, 600, 700, 850, 925 \\ 
        & & Atmospheric & u & 250, 500, 850\\ 
        & & Atmospheric & v & 250, 500, 850\\ 
        & & Atmospheric & t & 250, 500, 850\\ 
        & & Atmopheric & q & 250, 500, 600, 700, 850, 925                                                                            \\ \midrule
\multirow{8}{*}{AWI}     & \multirow{8}{*}{250km}               &  Single & t2m \\ 
        & & Single &  u10 \\ 
        & & Single & v10\\ 
        & & Atmospheric & z & 50, 250, 500, 600, 700, 850, 925\\ 
        & & Atmospheric & u & 50, 250, 500, 600, 700, 850, 925\\ 
        & & Atmospheric & v & 50, 250, 500, 600, 700, 850, 925\\
        & & Atmospheric & t & 50, 250, 500, 600, 700, 850, 925\\ 
        & & Atmospheric & q & 50, 250, 500, 600, 700, 850, 925\\ \midrule
\multirow{8}{*}{HAMMOZ}     & \multirow{8}{*}{250km}               &  Single & t2m \\ 
        & & Single &  u10 \\ 
        & & Single & v10\\ 
        & & Atmospheric & z & 50, 250, 500, 600, 700, 850, 925\\ 
        & & Atmospheric & u & 50, 250, 500, 600, 700, 850, 925\\ 
        & & Atmospheric & v & 50, 250, 500, 600, 700, 850, 925\\
        & & Atmospheric & t & 50, 250, 500, 600, 700, 850, 925\\ 
        & & Atmospheric & q & 50, 250, 500, 600, 700, 850, 925\\ \midrule
\multirow{4}{*}{CMCC}    & \multirow{4}{*}{100km}               &  Atmospheric & z & 50, 250, 500, 600, 700, 850, 925\\ 
& & Atmospheric & u & 50, 250, 500, 600, 700, 850, 925\\ 
& & Atmospheric & v & 50, 250, 500, 600, 700, 850, 925\\ 
& & Atmospheric & t & 250, 500, 850         \\
\bottomrule
\end{tabular}
\end{table}

We note that AWI and HAMMOZ are not the best data sources for higher resolution $1.40625\degree$ training, because their original resolution at $250$ km is lower than $1.40625\degree$, which is about $156$ km. We wanted to use other higher-resolution datasets but were not able to download them. We believe pretraining on other high-resolution datasets would lead to better performance.

\subsection{ERA5} \label{sec:app:era5_details}
We use the preprocessed version of ERA5 from WeatherBench~\citep{rasp2020weatherbench} for finetuning ClimaX. WeatherBench was created as a standard benchmark data and evaluation framework for comparing data-driven weather forecasting models. WeatherBench regridded the original ERA5 at $0.25\degree$ to three lower resolutions: $5.625\degree$, $2.8125\degree$, and $1.40625\degree$. See {\url{https://confluence.ecmwf.int/display/CKB/ERA5\%3A+data+documentation}} for more details of the raw ERA5 data.
Table~\ref{tab:era5_data} summarizes the variables we use for finetuning ClimaX.

\begin{table}[h]
    \centering
    \caption{{ECMWF variables used in our ERA5 dataset}. 
    \emph{Static} represents variables don't depend on time, \emph{Single}  represents surface variables, and \emph{Atmospheric} represents time-varying atmospheric properties at the chosen altitudes.}    
    \label{tab:era5_data}
    \begin{tabular}{rllll}
        \toprule
        Type & Variable name & Abbrev. & ECMWF ID & Levels
        \\
        \midrule
         Static & Land-sea mask & LSM & 172 & \\
         Static & Orography & & & \\
         Single & 2 metre temperature & T2m & 167 & \\
         Single & 10 metre U wind component & U10 & 165 & \\
         Single & 10 metre V wind component & V10 & 166 & \\
         \midrule
         Atmospheric & Geopotential & Z & 129 & 50, 250, 500, 600, 700, 850, 925 \\
         Atmospheric & U wind component & U & 131 & 50, 250, 500, 600, 700, 850, 925 \\
         Atmospheric & V wind component & V & 132 & 50, 250, 500, 600, 700, 850, 925 \\
         Atmospheric & Temperature & T & 130 & 50, 250, 500, 600, 700, 850, 925 \\
         Atmospheric & Specific humidity & Q & 133 & 50, 250, 500, 600, 700, 850, 925 \\
         Atmospheric & Relative humidity & R & 157 & 50, 250, 500, 600, 700, 850, 925 \\
        \bottomrule
    \end{tabular}
\end{table}

\subsubsection{ERA5-NA}
We constructed ERA5-NA from ERA5 to evaluate ClimaX and the baselines on regional forecasting. ERA-NA has the same set of variables as in Table~\ref{tab:era5_data}, but only contains data that belongs to the North America region. To do this, we first identified the latitude and longitude range to form a rectangular area that encapsulates North America, using the standard CORDEX domains \url{https://cordex.org/wp-content/uploads/2012/11/CORDEX-domain-description_231015.pdf}. For each data sample, we then extracted the spatial positions that fall into this range, forming in ERA5-NA.

\subsubsection{ERA-S2S}
We built ERA5-S2S from ERA5 to serve as a benchmark dataset for sub-seasonal to seasonal prediction. ERA5-S2S consists of two sub-datasets, whose the goals are to predict the biweekly average statistics of target variables in weeks $3$ and $4$, and weeks $5$ and $6$, respectively. The input includes all variables in Table~\ref{tab:era5_data}, while the output variables are are averaged over two weeks, starting from the start of week 3 (5) and to the end of week 4 (6).

\subsection{ClimateBench}
We refer to~\citet{watson2022climatebench} for complete details of ClimateBench.

\newpage
\section{Quantitative evaluation}

\subsection{Metrics}\label{sec:app:metrics}
This section presents all evaluation metrics we use in \Cref{sec:exps}. For all metrics, we denote $\Tilde{X}$ and $X$ as the prediction and ground truth, which have a shape of $N \times H \times W$, where $N$ is the number of forecasts, or the number of test samples, $H \times W$ is the spatial resolution. $L(i)$ is the latitude weighting term to account for the non-uniformity in areas of the grid cells. We have removed the time notation for simplicity.

\subsubsection{Weather forecasting metrics}
\paragraph{Root mean square error (RMSE)}
\begin{equation}
    \text{RMSE} = \frac{1}{N} \sum_{k=1}^{N} \sqrt{\frac{1}{H \times W} \sum_{i=1}^H \sum_{j=1}^W L(i)(\Tilde{X}_{k,i,j} - X_{k,i,j})^2}. \label{eq:lat_rmse}
\end{equation}

\paragraph{Anomaly correlation coefficient (ACC)}
Anomaly correlation coefficient (ACC) is the spatial correlation between prediction anomalies $\Tilde{X}^{'}$ relative to climatology and ground truth anomalies $X^{'}$ relative to climatology:
\begin{gather}
    \text{ACC} = \frac{\sum_{k,i,j} L(i) \Tilde{X}^{'}_{k,i,j} X^{'}_{k,i,j}}{\sqrt{\sum_{k,i,j} L(i) \Tilde{X}^{'2}_{k,i,j} \sum_{k,i,j} L(i) X^{'2}_{k,i,j}}}, \\
    \Tilde{X}^{'} = \Tilde{X}^{'} - C, X^{'} = X^{'} - C,
\end{gather}
in which climatology $C$ is the temporal mean of the ground truth data over the entire test set $C = \frac{1}{N}\sum_k X$.

\subsubsection{Climate projection metrics}
\paragraph{Normalized spatial root mean square error (NRMSE$_s$)}
Normalized spatial root mean square error (NRMSE$_s$) measures the spatial discrepancy between the temporal mean of the prediction and the temporal mean of the ground truth:
\begin{equation}
    \text{NRMSE}_s = \sqrt{\left\langle \left(\frac{1}{N} \sum_{k=1}^N \Tilde{X} - \frac{1}{N} \sum_{k=1}^N X \right)^2 \right\rangle} \bigg/ \frac{1}{N} \sum_{k=1}^N \left\langle X \right\rangle,
\end{equation}
in which $\langle A \rangle$ is the global mean of $A$:
\begin{equation}
    \langle A \rangle = \frac{1}{H \times W} \sum_{i=1}^H \sum_{j=1}^W L(i) A_{i,j}
\end{equation}

\paragraph{Normalized global root mean square error (NRMSE$_g$) }
Normalized global root mean square error (NRMSE$_g$) measures the discrepancy between the global mean of the prediction and the global mean of the ground truth:
\begin{equation}
    \text{NRMSE}_g = \sqrt{\frac{1}{N} \sum_{k=1}^N \left(\langle \Tilde{X} \rangle - \langle X \rangle\right)^2}  \bigg/ \frac{1}{N} \sum_{k=1}^N \left\langle X \right\rangle.
\end{equation}

\paragraph{Total normalized root mean square error (TRMSE)}
Total normalized root mean square error (TRMSE) is the weighted sum of NRMSE$_s$ and NRMSE$_g$:
\begin{equation}
    \text{TRMSE} = \text{NRMSE}_s + \alpha \cdot \text{NRMSE}_g,
\end{equation}
where $\alpha$ is chosen to be $5$ as suggested by~\citet{watson2022climatebench}.

\subsubsection{Climate downscaling metrics}
\paragraph{Root mean square error (RMSE)}
This is the same as \Cref{eq:lat_rmse}. 

\paragraph{Mean bias}
Mean bias measures the difference between the spatial mean of the prediction and the spatial mean of the ground truth. A positive mean bias shows an overestimation, while a negative mean bias shows an underestimation of the mean value.
\begin{equation}
    \text{Mean bias} = \frac{1}{N \times H \times W} \sum_{k=1}^N \sum_{i=1}^H \sum_{j=1}^W \Tilde{X} - \frac{1}{N \times H \times W} \sum_{k=1}^N \sum_{i=1}^H \sum_{j=1}^W X
\end{equation}

\paragraph{Pearson coefficient}
Pearson coefficient measures the correlation between the prediction and the ground truth. We first flatten the prediction and ground truth, and compute the metric as follows:
\begin{equation}
    \rho_{\Tilde{X}, X}=\frac{\operatorname{cov}(\Tilde{X}, X)}{\sigma_{\Tilde{X}} \sigma_X}
\end{equation}

\subsection{Results summary} \label{app:results_summary}
Table~\ref{tab:rmse-compare} and~\ref{tab:acc-compare} summarize the global forecasting results of ClimaX and the baselines for all target variables and at all lead times. In addition to IFS and the two CNN-based baselines in the main text, we include FourCastNet~\citep{pathak2022fourcastnet}, PanguWeather~\citep{bi2022pangu}, and GraphCast~\citep{lam2022graphcast} for comprehensiveness. We want to emphasize that the results obtained by these methods are not comparable with ClimaX, as they were trained on ERA5 at $0.25\degree$, a much higher resolution compared to $5.625\degree$ and $1.40625\degree$ data used to train ClimaX. In \Cref{sec:scaling}, we had a discussion on how the performance of ClimaX scales favorably with respect to data resolution. We hope this summary will provide future works with an easier comparison with existing baselines.

In spite of being trained on much lower resolutions, ClimaX outperforms FourCastNet in forecasting Z500, T850, and U10 at lead times from 3 days and beyond, in terms of both RMSE and ACC. For T2m, ClimaX achieves better results at horizons longer than 3 days. PanguWeather performs better than ClimaX on most of the tasks, but the gap between the two methods shrinks and becomes negligible as the lead time increases. ClimaX even outperforms PanguWeather in predicting U10 at 7 days lead times. This is because ClimaX is finetuned to perform direct prediction, which mitigates error accumulation for long horizon prediction. GraphCast achieves the lowest RMSE among all methods, but performs worse in terms of ACC compared to ClimaX and PanguWeather.

\begin{table}[t]
    \centering
    \caption{RMSE on global forecasting for different target variables at different lead times. Lower is better.}
    \resizebox{\textwidth}{!}{\begin{NiceTabular}{@{}llrrrrrrrrrr@{}}
    \toprule
        \multirow{2}{*}{\textsc{Variable}} & {\textsc{Lead time}} &\multicolumn{2}{c}{ClimaX} & \multicolumn{1}{c}{FCN\tabularnote{FourCastNet~\citep{pathak2022fourcastnet}}} & \multicolumn{1}{c}{PW\tabularnote{PanguWeather~\citep{bi2022pangu}}} & \multicolumn{1}{c}{GC\tabularnote{GraphCast~\citep{lam2022graphcast}}} & \multicolumn{1}{c}{HRES} & \multicolumn{2}{c}{IFS} & \multicolumn{1}{c}{ResNet} & \multicolumn{1}{c}{UNet} \\
        \cmidrule(lr){3-4} \cmidrule(lr){5-5}\cmidrule(lr){6-6}\cmidrule(lr){7-7}\cmidrule(lr){8-8}\cmidrule(lr){9-10}\cmidrule(lr){11-11}\cmidrule(lr){12-12}
        {} & {[hr.]}& {$5.625\degree$} & {$1.40625\degree$} & {$0.25\degree$} & {$0.25\degree$} & {$0.25\degree$} & {0.1} & {$5.625\degree$} & {$1.40625\degree$} & {$5.625\degree$} & {$5.625\degree$}\\
        \midrule
Z500 & 6 & \rmsegradientz{62.73} & \rmsegradientz{49.67} & \rmsegradientz{37.52} & \rmsegradientz{15.40} & \rmsegradientz{16.46} & \rmsegradientz{24.66} & \rmsegradientz{26.93} & \rmsegradientz{26.96} & \rmsegradientz{47.00} & \rmsegradientz{53.66} \\
  [m$^2$/s$^2$]& 24 & \rmsegradientz{96.19} & \rmsegradientz{72.76} & \rmsegradientz{81.31} & \rmsegradientz{42.23} & \rmsegradientz{38.77} & \rmsegradientz{45.90} & \rmsegradientz{51.01} & \rmsegradientz{50.96} & \rmsegradientz{86.60} & \rmsegradientz{132.65} \\
 & 72 & \rmsegradientz{244.08} & \rmsegradientz{201.00} & \rmsegradientz{251.96} & \rmsegradientz{133.12} & \rmsegradientz{125.78} & \rmsegradientz{146.37} & \rmsegradientz{152.15} & \rmsegradientz{152.20} & \rmsegradientz{305.22} & \rmsegradientz{458.84} \\
 & 120 & \rmsegradientz{440.40} & \rmsegradientz{392.00} & \rmsegradientz{483.44} & \rmsegradientz{295.63} & \rmsegradientz{271.65} & \rmsegradientz{316.79} & \rmsegradientz{331.45} & \rmsegradientz{331.38} & \rmsegradientz{614.20} & \rmsegradientz{721.83} \\
 & 168 & \rmsegradientz{599.43} & \rmsegradientz{566.00} & \rmsegradientz{680.00} & \rmsegradientz{504.90} & \rmsegradientz{466.53} & \rmsegradientz{535.93} & \rmsegradientz{549.01} & \rmsegradientz{548.96} & \rmsegradientz{806.59} & \rmsegradientz{819.39} \\
 & 336 & \rmsegradientz{790.26} & \rmsegradientz{788.43} & nan & nan & nan & nan & \rmsegradientz{1011.72} & \rmsegradientz{1011.56} & \rmsegradientz{835.55} & \rmsegradientz{866.40} \\
 & 720 & \rmsegradientz{815.25} & \rmsegradientz{817.52} & nan & nan & nan & nan & nan & nan & \rmsegradientz{858.98} & \rmsegradientz{880.34} \\
\midrule
T2m & 6 & \rmsegradientt{0.95} & \rmsegradientt{1.11} & \rmsegradientt{0.72} & \rmsegradientt{0.59} & \rmsegradientt{0.50} & \rmsegradientt{0.35} & \rmsegradientt{0.97} & \rmsegradientt{0.97} & \rmsegradientt{0.76} & \rmsegradientt{0.77} \\
 [K]& 24 & \rmsegradientt{1.10} & \rmsegradientt{1.19} & \rmsegradientt{0.95} & \rmsegradientt{0.72} & \rmsegradientt{0.62} & \rmsegradientt{0.66} & \rmsegradientt{1.02} & \rmsegradientt{1.02} & \rmsegradientt{0.91} & \rmsegradientt{1.11} \\
 & 72 & \rmsegradientt{1.43} & \rmsegradientt{1.47} & \rmsegradientt{1.38} & \rmsegradientt{1.05} & \rmsegradientt{0.94} & \rmsegradientt{1.06} & \rmsegradientt{1.30} & \rmsegradientt{1.30} & \rmsegradientt{1.70} & \rmsegradientt{1.91} \\
 & 120 & \rmsegradientt{1.83} & \rmsegradientt{1.83} & \rmsegradientt{1.99} & \rmsegradientt{1.53} & \rmsegradientt{1.36} & \rmsegradientt{1.52} & \rmsegradientt{1.72} & \rmsegradientt{1.71} & \rmsegradientt{2.22} & \rmsegradientt{2.49} \\
 & 168 & \rmsegradientt{2.18} & \rmsegradientt{2.17} & \rmsegradientt{2.54} & \rmsegradientt{2.06} & \rmsegradientt{1.88} & \rmsegradientt{2.06} & \rmsegradientt{2.24} & \rmsegradientt{2.23} & \rmsegradientt{2.66} & \rmsegradientt{2.66} \\
 & 336 & \rmsegradientt{2.61} & \rmsegradientt{2.67} & nan & nan & nan & nan & \rmsegradientt{3.31} & \rmsegradientt{3.30} & \rmsegradientt{2.86} & \rmsegradientt{2.79} \\
 & 720 & \rmsegradientt{2.67} & \rmsegradientt{2.74} & nan & nan & nan & nan & nan & nan & \rmsegradientt{2.86} & \rmsegradientt{2.81} \\
\midrule
T850 & 6 & \rmsegradientt{0.88} & \rmsegradientt{0.84} & \rmsegradientt{0.52} & \rmsegradientt{0.42} & \rmsegradientt{0.28} & \rmsegradientt{0.33} & \rmsegradientt{0.69} & \rmsegradientt{0.69} & \rmsegradientt{0.70} & \rmsegradientt{0.80} \\
 [K]& 24 & \rmsegradientt{1.11} & \rmsegradientt{1.02} & \rmsegradientt{0.81} & \rmsegradientt{0.72} & \rmsegradientt{0.58} & \rmsegradientt{0.70} & \rmsegradientt{0.87} & \rmsegradientt{0.87} & \rmsegradientt{1.26} & \rmsegradientt{1.25} \\
 & 72 & \rmsegradientt{1.59} & \rmsegradientt{1.46} & \rmsegradientt{1.55} & \rmsegradientt{1.13} & \rmsegradientt{1.02} & \rmsegradientt{1.27} & \rmsegradientt{1.34} & \rmsegradientt{1.34} & \rmsegradientt{1.90} & \rmsegradientt{2.39} \\
 & 120 & \rmsegradientt{2.23} & \rmsegradientt{2.08} & \rmsegradientt{2.47} & \rmsegradientt{1.78} & \rmsegradientt{1.63} & \rmsegradientt{1.96} & \rmsegradientt{2.01} & \rmsegradientt{2.01} & \rmsegradientt{2.86} & \rmsegradientt{3.23} \\
 & 168 & \rmsegradientt{2.77} & \rmsegradientt{2.66} & \rmsegradientt{3.30} & \rmsegradientt{2.60} & \rmsegradientt{2.41} & \rmsegradientt{2.78} & \rmsegradientt{2.82} & \rmsegradientt{2.82} & \rmsegradientt{3.51} & \rmsegradientt{3.50} \\
 & 336 & \rmsegradientt{3.40} & \rmsegradientt{3.41} & nan & nan & nan & nan & \rmsegradientt{4.43} & \rmsegradientt{4.43} & \rmsegradientt{3.65} & \rmsegradientt{3.65} \\
 & 720 & \rmsegradientt{3.47} & \rmsegradientt{3.49} & nan & nan & nan & nan & nan & nan & \rmsegradientt{3.69} & \rmsegradientt{3.73} \\
\midrule
U10 & 6 & \rmsegradientu{1.08} & \rmsegradientu{1.04} & \rmsegradientu{0.55} & \rmsegradientu{0.46} & \rmsegradientu{0.37} & \rmsegradientu{0.58} & \rmsegradientu{0.80} & \rmsegradientu{0.79} & \rmsegradientu{0.86} & \rmsegradientu{1.02} \\
 [m/s]& 24 & \rmsegradientu{1.41} & \rmsegradientu{1.31} & \rmsegradientu{0.99} & \rmsegradientu{0.90} & \rmsegradientu{0.80} & \rmsegradientu{1.15} & \rmsegradientu{1.11} & \rmsegradientu{1.11} & \rmsegradientu{1.27} & \rmsegradientu{1.68} \\
 & 72 & \rmsegradientu{2.18} & \rmsegradientu{2.02} & \rmsegradientu{2.24} & \rmsegradientu{1.60} & \rmsegradientu{1.47} & \rmsegradientu{1.98} & \rmsegradientu{1.92} & \rmsegradientu{1.92} & \rmsegradientu{2.78} & \rmsegradientu{3.17} \\
 & 120 & \rmsegradientu{2.94} & \rmsegradientu{2.79} & \rmsegradientu{3.41} & \rmsegradientu{2.52} & \rmsegradientu{2.36} & \rmsegradientu{2.95} & \rmsegradientu{2.89} & \rmsegradientu{2.89} & \rmsegradientu{3.63} & \rmsegradientu{3.93} \\
 & 168 & \rmsegradientu{3.43} & \rmsegradientu{3.35} & \rmsegradientu{4.18} & \rmsegradientu{3.46} & \rmsegradientu{3.25} & \rmsegradientu{3.87} & \rmsegradientu{3.81} & \rmsegradientu{3.81} & \rmsegradientu{4.15} & \rmsegradientu{4.08} \\
 & 336 & \rmsegradientu{3.91} & \rmsegradientu{3.92} & nan & nan & nan & nan & \rmsegradientu{5.24} & \rmsegradientu{5.23} & \rmsegradientu{4.23} & \rmsegradientu{4.16} \\
 & 720 & \rmsegradientu{3.96} & \rmsegradientu{3.97} & nan & nan & nan & nan & nan & nan & \rmsegradientu{4.29} & \rmsegradientu{4.22} \\
         \bottomrule
    \end{NiceTabular}}
    \label{tab:rmse-compare}
\end{table}

\begin{table}[t]
    \centering
    \caption{ACC on global forecasting for different target variables at different lead times. Higher is better.}
    \label{tab:acc-compare}
   \resizebox{\textwidth}{!}{\begin{NiceTabular}{@{}llrrrrrrrrrr@{}}
    \toprule
        \multirow{2}{*}{\textsc{Variable}} & {\textsc{Lead time}} &\multicolumn{2}{c}{ClimaX} & \multicolumn{1}{c}{FCN\tabularnote{FourCastNet~\citep{pathak2022fourcastnet}}} & \multicolumn{1}{c}{PW\tabularnote{PanguWeather~\citep{bi2022pangu}}} & \multicolumn{1}{c}{GC\tabularnote{GraphCast~\citep{lam2022graphcast}}} & \multicolumn{1}{c}{HRES} & \multicolumn{2}{c}{IFS} & \multicolumn{1}{c}{ResNet} & \multicolumn{1}{c}{UNet} \\
        \cmidrule(lr){3-4} \cmidrule(lr){5-5}\cmidrule(lr){6-6}\cmidrule(lr){7-7}\cmidrule(lr){8-8}\cmidrule(lr){9-10}\cmidrule(lr){11-11}\cmidrule(lr){12-12}
        {} & {[hr.]}& {$5.625\degree$} & {$1.40625\degree$} & {$0.25\degree$} & {$0.25\degree$} & {$0.25\degree$} & {0.1} & {$5.625\degree$} & {$1.40625\degree$} & {$5.625\degree$} & {$5.625\degree$}\\
        \midrule
Z500 & 6 & \accgradient{1.00} & \accgradient{1.00} & \accgradient{1.00} & \accgradient{1.00} & \accgradient{1.00} & \accgradient{1.00} & \accgradient{1.00} & \accgradient{1.00} & \accgradient{1.00} & \accgradient{1.00} \\
 & 24 & \accgradient{1.00} & \accgradient{1.00} & \accgradient{1.00} & \accgradient{1.00} & \accgradient{1.00} & \accgradient{1.00} & \accgradient{1.00} & \accgradient{1.00} & \accgradient{1.00} & \accgradient{0.99} \\
 & 72 & \accgradient{0.97} & \accgradient{0.98} & \accgradient{0.97} & \accgradient{0.99} & \accgradient{0.99} & \accgradient{0.98} & \accgradient{0.99} & \accgradient{0.99} & \accgradient{0.95} & \accgradient{0.89} \\
 & 120 & \accgradient{0.90} & \accgradient{0.92} & \accgradient{0.89} & \accgradient{0.96} & \accgradient{0.94} & \accgradient{0.92} & \accgradient{0.95} & \accgradient{0.95} & \accgradient{0.79} & \accgradient{0.69} \\
 & 168 & \accgradient{0.80} & \accgradient{0.82} & \accgradient{0.76} & \accgradient{0.87} & \accgradient{0.83} & \accgradient{0.78} & \accgradient{0.87} & \accgradient{0.87} & \accgradient{0.57} & \accgradient{0.57} \\
 & 336 & \accgradient{0.59} & \accgradient{0.59} & nan & nan & nan & nan & \accgradient{0.55} & \accgradient{0.55} & \accgradient{0.53} & \accgradient{0.51} \\
 & 720 & \accgradient{0.55} & \accgradient{0.55} & nan & nan & nan & nan & nan & nan & \accgradient{0.49} & \accgradient{0.49} \\
\midrule
T2m & 6 & \accgradient{0.98} & \accgradient{0.98} & \accgradient{0.99} & \accgradient{0.99} & \accgradient{0.98} & \accgradient{0.99} & \accgradient{0.99} & \accgradient{0.99} & \accgradient{0.99} & \accgradient{0.99} \\
  & 24 & \accgradient{0.98} & \accgradient{0.97} & \accgradient{0.98} & \accgradient{0.99} & \accgradient{0.98} & \accgradient{0.98} & \accgradient{0.99} & \accgradient{0.99} & \accgradient{0.98} & \accgradient{0.98} \\
 & 72 & \accgradient{0.96} & \accgradient{0.96} & \accgradient{0.96} & \accgradient{0.98} & \accgradient{0.95} & \accgradient{0.94} & \accgradient{0.98} & \accgradient{0.98} & \accgradient{0.94} & \accgradient{0.93} \\
 & 120 & \accgradient{0.94} & \accgradient{0.94} & \accgradient{0.92} & \accgradient{0.95} & \accgradient{0.90} & \accgradient{0.88} & \accgradient{0.96} & \accgradient{0.96} & \accgradient{0.90} & \accgradient{0.88} \\
 & 168 & \accgradient{0.91} & \accgradient{0.91} & \accgradient{0.87} & \accgradient{0.92} & \accgradient{0.81} & \accgradient{0.77} & \accgradient{0.93} & \accgradient{0.93} & \accgradient{0.86} & \accgradient{0.86} \\
 & 336 & \accgradient{0.86} & \accgradient{0.85} & nan & nan & nan & nan & \accgradient{0.85} & \accgradient{0.85} & \accgradient{0.83} & \accgradient{0.84} \\
 & 720 & \accgradient{0.85} & \accgradient{0.84} & nan & nan & nan & nan & nan & nan & \accgradient{0.83} & \accgradient{0.83} \\
\midrule
T850 & 6 & \accgradient{0.98} & \accgradient{0.99} & \accgradient{0.99} & \accgradient{1.00} & \accgradient{1.00} & \accgradient{0.99} & \accgradient{0.99} & \accgradient{0.99} & \accgradient{0.99} & \accgradient{0.99} \\
 & 24 & \accgradient{0.98} & \accgradient{0.98} & \accgradient{0.98} & \accgradient{0.99} & \accgradient{0.99} & \accgradient{0.98} & \accgradient{0.99} & \accgradient{0.99} & \accgradient{0.97} & \accgradient{0.97} \\
 & 72 & \accgradient{0.95} & \accgradient{0.96} & \accgradient{0.95} & \accgradient{0.98} & \accgradient{0.96} & \accgradient{0.93} & \accgradient{0.97} & \accgradient{0.97} & \accgradient{0.92} & \accgradient{0.88} \\
 & 120 & \accgradient{0.89} & \accgradient{0.91} & \accgradient{0.87} & \accgradient{0.94} & \accgradient{0.89} & \accgradient{0.84} & \accgradient{0.93} & \accgradient{0.94} & \accgradient{0.82} & \accgradient{0.75} \\
 & 168 & \accgradient{0.82} & \accgradient{0.84} & \accgradient{0.77} & \accgradient{0.87} & \accgradient{0.75} & \accgradient{0.68} & \accgradient{0.87} & \accgradient{0.87} & \accgradient{0.68} & \accgradient{0.69} \\
 & 336 & \accgradient{0.71} & \accgradient{0.71} & nan & nan & nan & nan & \accgradient{0.68} & \accgradient{0.69} & \accgradient{0.66} & \accgradient{0.66} \\
 & 720 & \accgradient{0.69} & \accgradient{0.68} & nan & nan & nan & nan & nan & nan & \accgradient{0.64} & \accgradient{0.64} \\
\midrule
U10 & 6 & \accgradient{0.97} & \accgradient{0.97} & \accgradient{0.99} & \accgradient{0.99} & \accgradient{0.99} & \accgradient{0.99} & \accgradient{0.98} & \accgradient{0.98} & \accgradient{0.98} & \accgradient{0.97} \\
  & 24 & \accgradient{0.94} & \accgradient{0.95} & \accgradient{0.97} & \accgradient{0.97} & \accgradient{0.98} & \accgradient{0.96} & \accgradient{0.97} & \accgradient{0.97} & \accgradient{0.95} & \accgradient{0.91} \\
 & 72 & \accgradient{0.85} & \accgradient{0.87} & \accgradient{0.85} & \accgradient{0.92} & \accgradient{0.93} & \accgradient{0.88} & \accgradient{0.89} & \accgradient{0.89} & \accgradient{0.74} & \accgradient{0.65} \\
 & 120 & \accgradient{0.70} & \accgradient{0.74} & \accgradient{0.64} & \accgradient{0.80} & \accgradient{0.82} & \accgradient{0.74} & \accgradient{0.76} & \accgradient{0.76} & \accgradient{0.52} & \accgradient{0.37} \\
 & 168 & \accgradient{0.56} & \accgradient{0.59} & \accgradient{0.45} & \accgradient{0.63} & \accgradient{0.64} & \accgradient{0.55} & \accgradient{0.58} & \accgradient{0.58} & \accgradient{0.28} & \accgradient{0.28} \\
 & 336 & \accgradient{0.33} & \accgradient{0.32} & nan & nan & nan & nan & \accgradient{0.21} & \accgradient{0.21} & \accgradient{0.19} & \accgradient{0.22} \\
 & 720 & \accgradient{0.29} & \accgradient{0.28} & nan & nan & nan & nan & nan & nan & \accgradient{0.17} & \accgradient{0.21} \\
         \bottomrule
    \end{NiceTabular}}
\end{table}

\clearpage
\section{Qualitative evaluation}\label{appendix:qualeval}
We qualitatively evaluate the performance of CliMax on global forecasting tasks for all target variables and at all lead times. In each figure, the first column is the initial condition of the target variable, which serves as the input, the second column is the ground truth of the target variable at a particular lead time, the third column is the prediction of ClimaX, and the last column is the bias, which is the difference between the prediction and the ground truth.

\subsection{Nowcasting}
\begin{figure}[h!]
    \centering
    \begin{subfigure}[b]{\textwidth}
        \includegraphics[width=\textwidth]{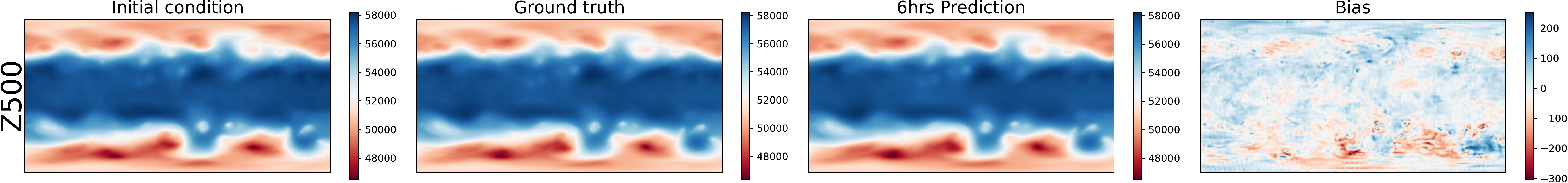}
        \includegraphics[width=\textwidth]{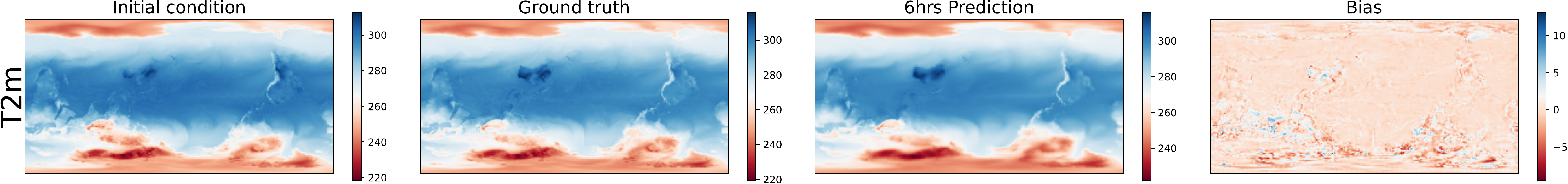}
        \includegraphics[width=\textwidth]{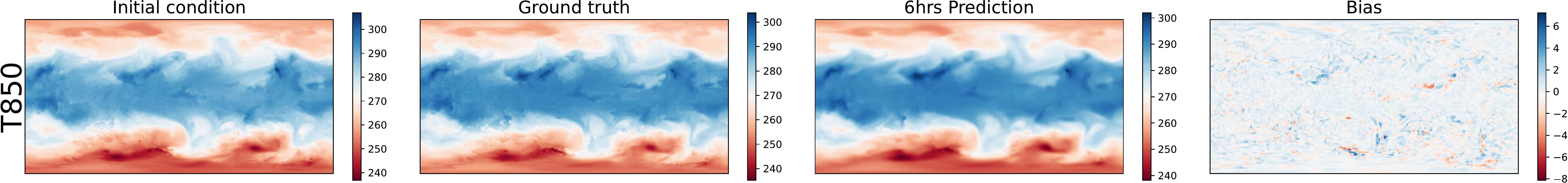}
        \includegraphics[width=\textwidth]{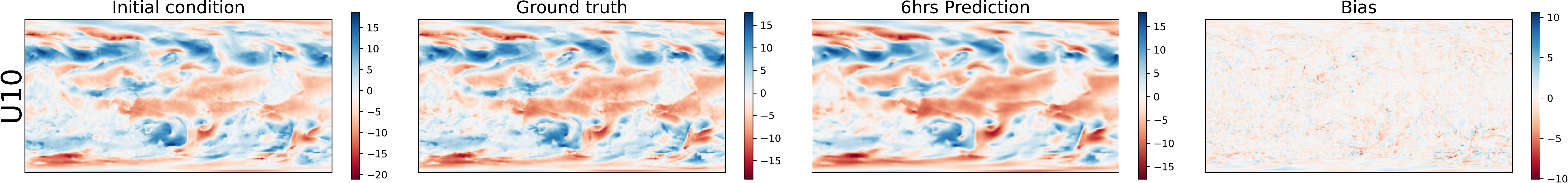}
    \end{subfigure}
    \caption{Example forecasts from ClimaX at 6-hour lead time compared to ground truth ERA5.}
    \label{fig:6-hrs-qual}
\end{figure}

\newpage
\subsection{Short and medium-range weather forecasting}

\begin{figure}[h!]
    \centering
    \begin{subfigure}[b]{\textwidth}
        \includegraphics[width=\textwidth]{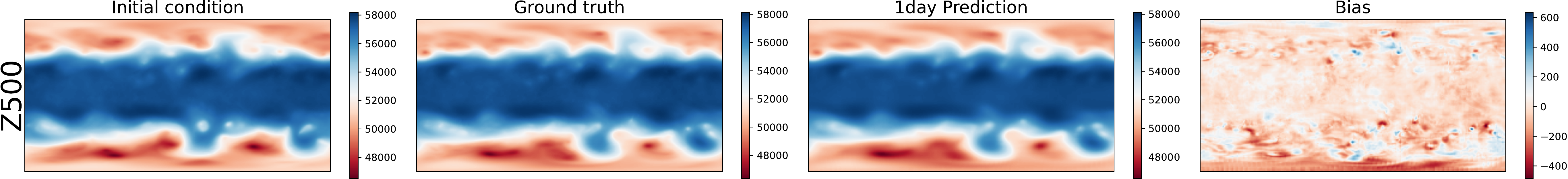}
        \includegraphics[width=\textwidth]{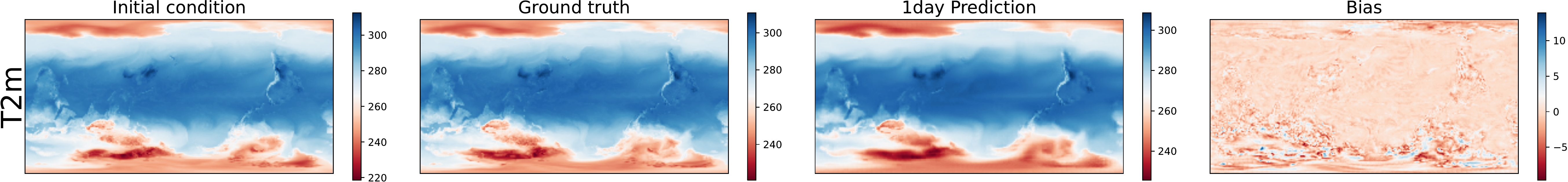}
        \includegraphics[width=\textwidth]{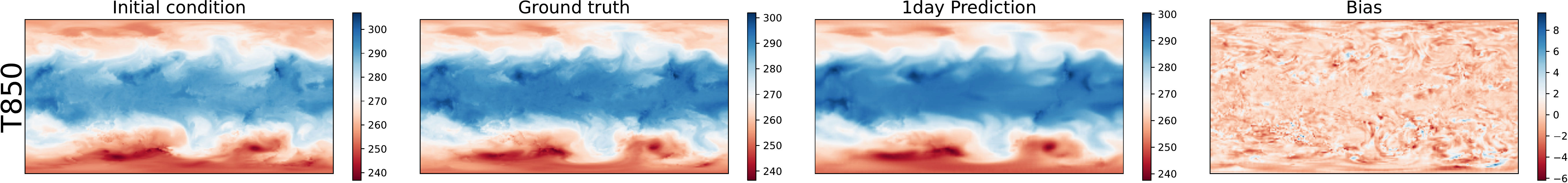}
        \includegraphics[width=\textwidth]{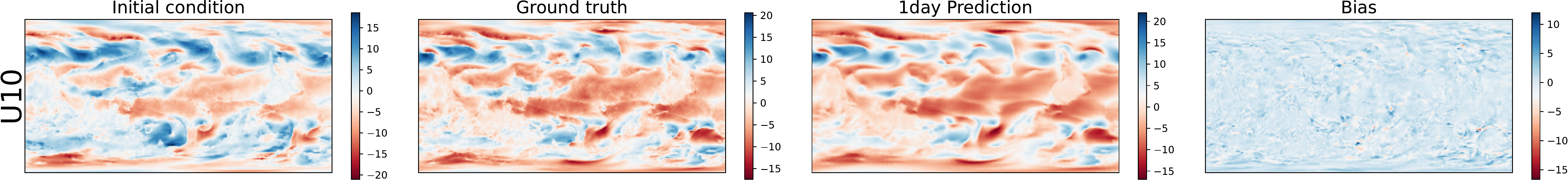}
    \end{subfigure}
    \caption{Example forecasts from ClimaX at 1-day lead time compared to ground truth ERA5.}
    \label{fig:1-day-qual}
\end{figure}
\begin{figure}[h!]
    \centering
    \begin{subfigure}[b]{\textwidth}
        \includegraphics[width=\textwidth]{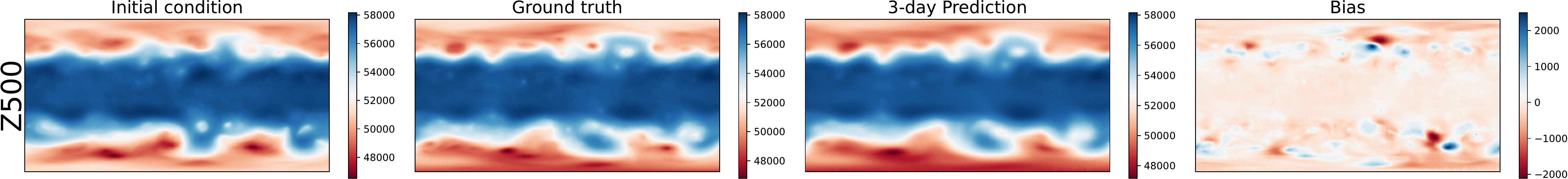}
        \includegraphics[width=\textwidth]{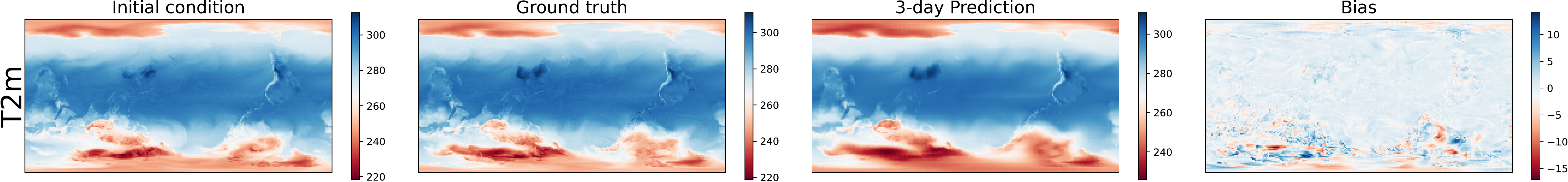}
        \includegraphics[width=\textwidth]{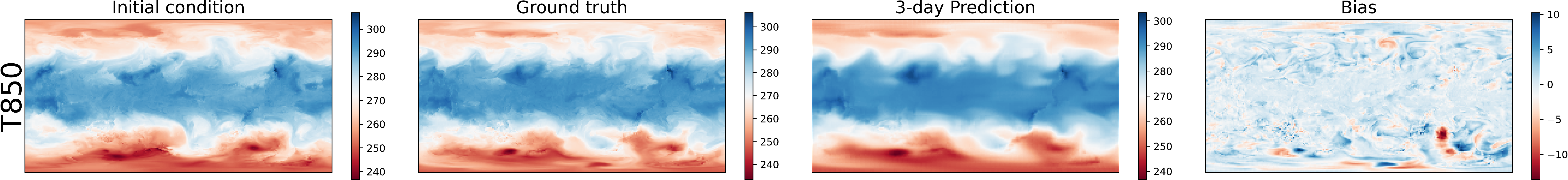}
        \includegraphics[width=\textwidth]{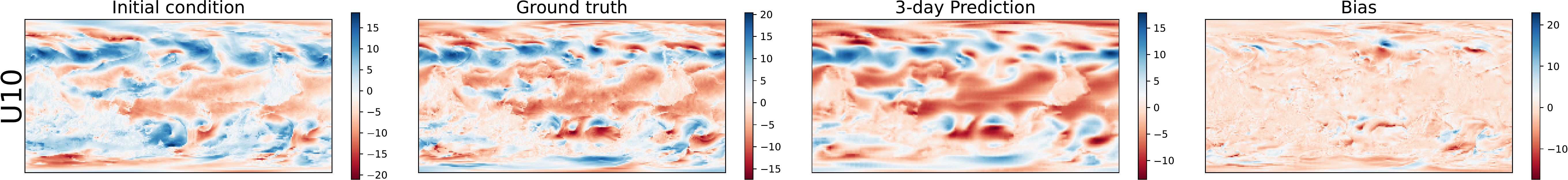}
    \end{subfigure}
    \caption{Example forecasts from ClimaX at 3-day lead time compared to ground truth ERA5.}
    \label{fig:3-day-qual}
\end{figure}

\begin{figure}[h!]
    \centering
    \begin{subfigure}[b]{\textwidth}
        \includegraphics[width=\textwidth]{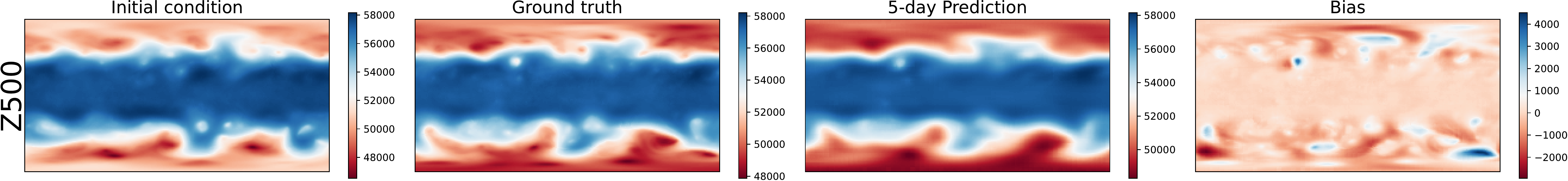}
        \includegraphics[width=\textwidth]{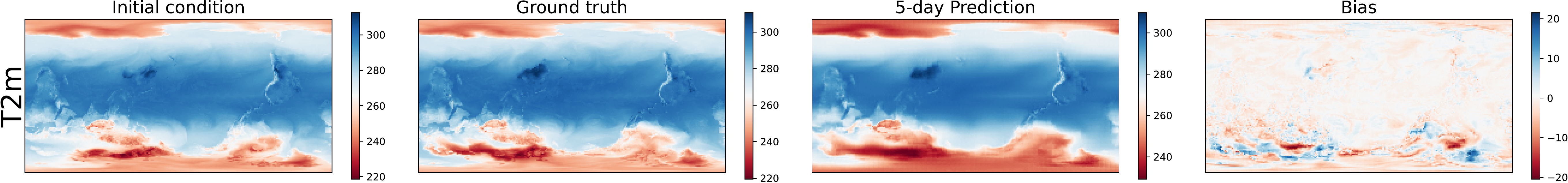}
        \includegraphics[width=\textwidth]{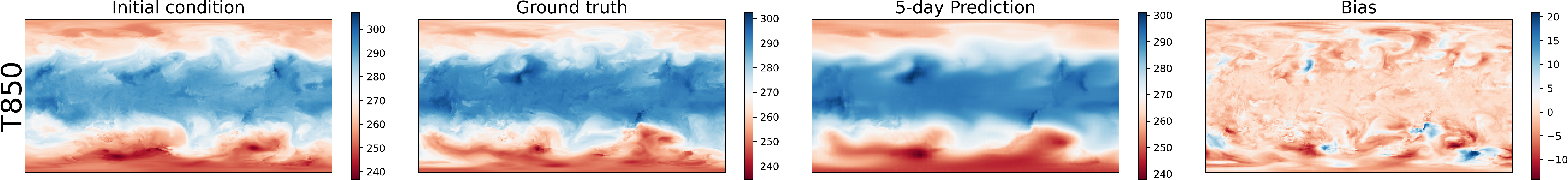}
        \includegraphics[width=\textwidth]{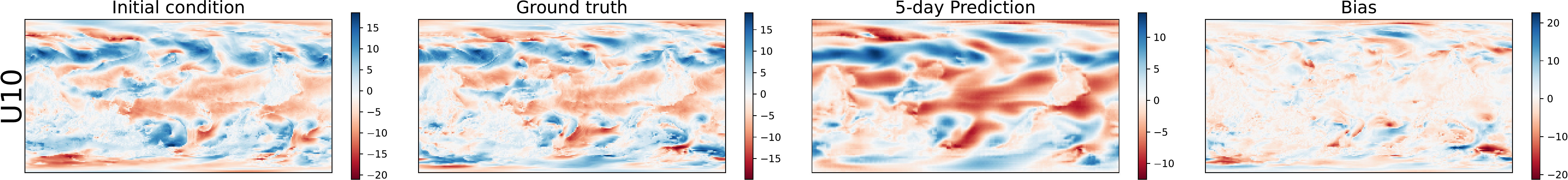}
    \end{subfigure}
    \caption{Example forecasts from ClimaX at 5-day lead time compared to ground truth ERA5.}
    \label{fig:5-day-qual}
\end{figure}
\begin{figure}[h!]
    \centering
    \begin{subfigure}[b]{\textwidth}
        \includegraphics[width=\textwidth]{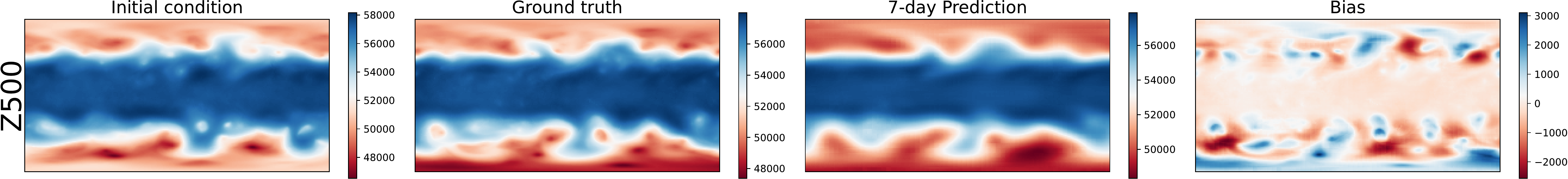}
        \includegraphics[width=\textwidth]{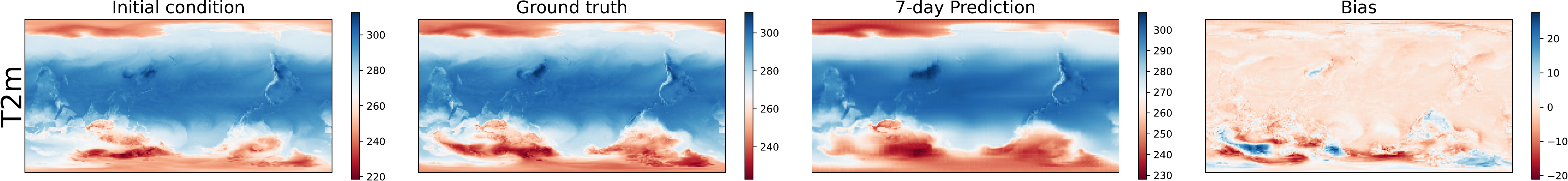}
        \includegraphics[width=\textwidth]{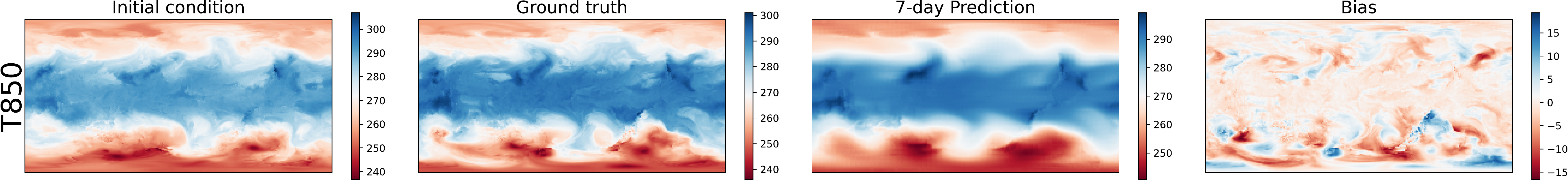}
        \includegraphics[width=\textwidth]{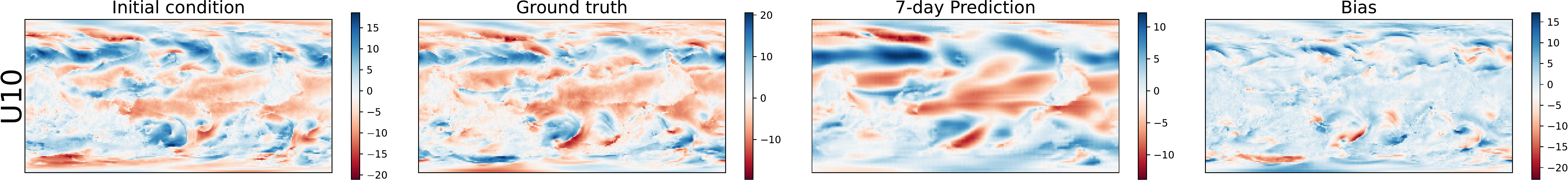}
    \end{subfigure}
    \caption{Example forecasts from ClimaX at 7-day lead time compared to ground truth ERA5.}
    \label{fig:7-day-qual}
\end{figure}

\clearpage

\subsection{Longer horizon instantaneous forecasting}

\begin{figure}[h!]
    \centering
    \begin{subfigure}[b]{\textwidth}
        \includegraphics[width=\textwidth]{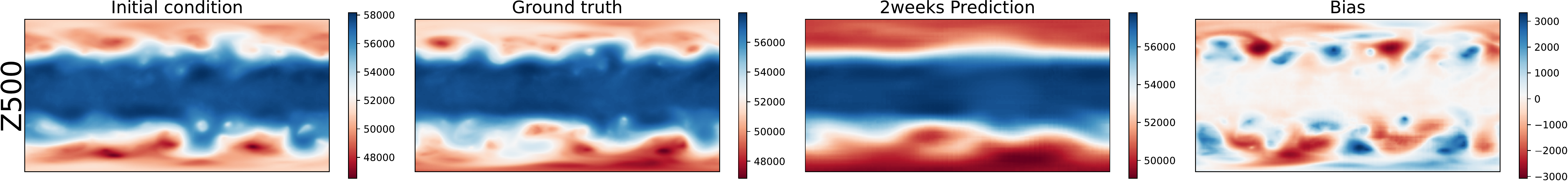}
        \includegraphics[width=\textwidth]{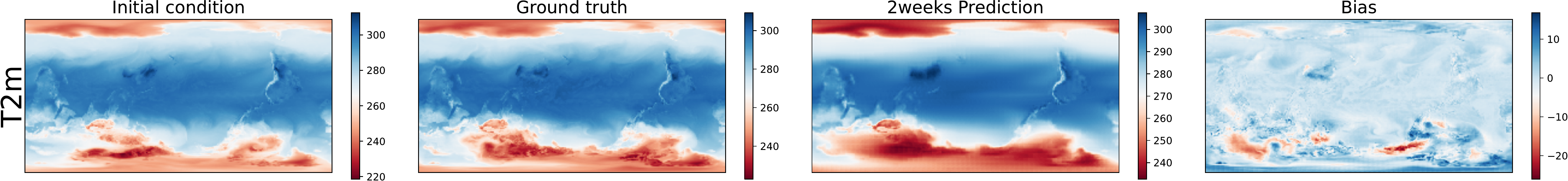}
        \includegraphics[width=\textwidth]{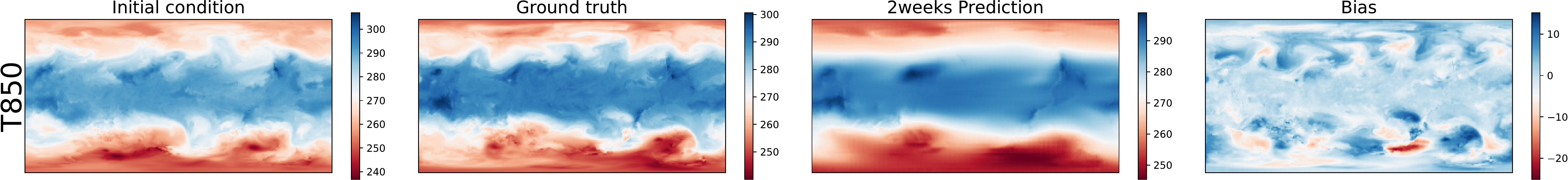}
        \includegraphics[width=\textwidth]{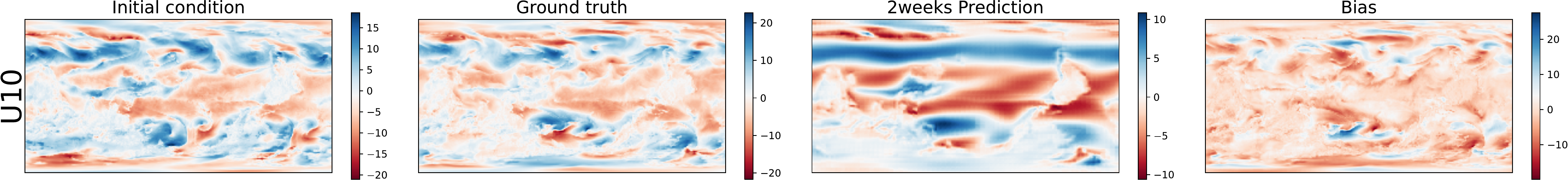}
    \end{subfigure}
    \caption{Example forecasts from ClimaX at 2-week lead time compared to ground truth ERA5.}
    \label{fig:2-weeks-qual}
\end{figure}
\begin{figure}[h!]
    \centering
    \begin{subfigure}[b]{\textwidth}
        \includegraphics[width=\textwidth]{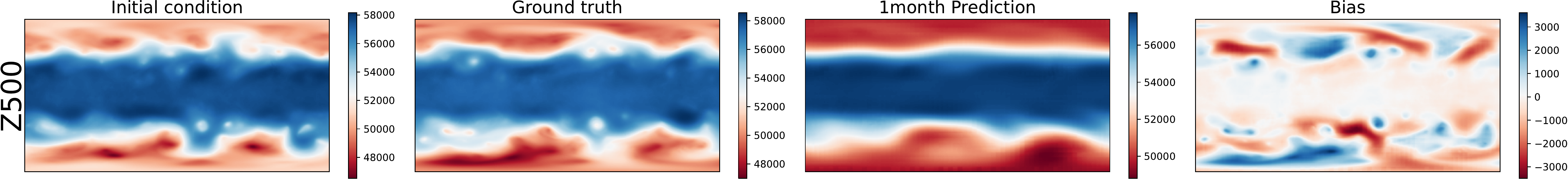}
        \includegraphics[width=\textwidth]{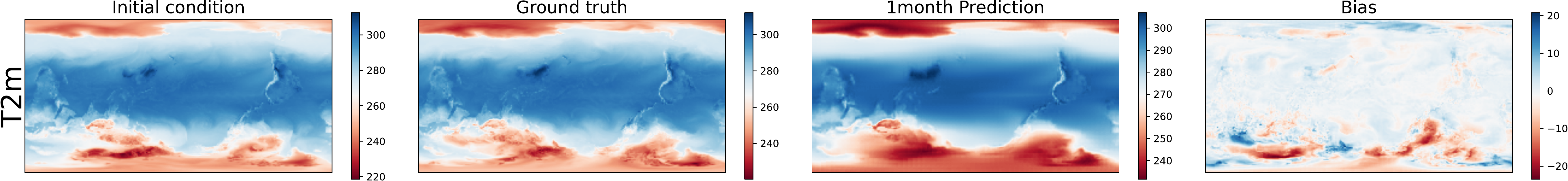}
        \includegraphics[width=\textwidth]{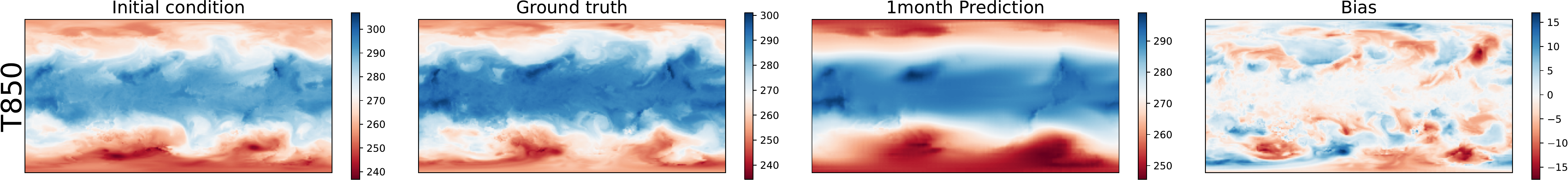}
        \includegraphics[width=\textwidth]{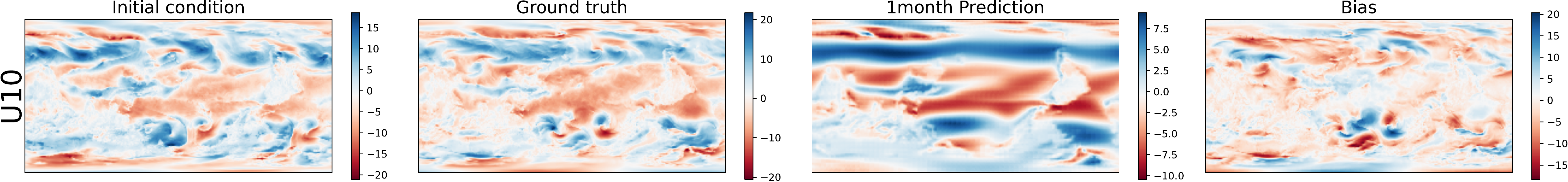}
    \end{subfigure}
    \caption{Example forecasts from ClimaX at  1-month lead time compared to ground truth ERA5.}
    \label{fig:1-month-qual}
\end{figure}

\end{document}